\PassOptionsToPackage{usenames,dvipsnames}{xcolor}
\documentclass[sigconf]{acmart}%\documentclass[sigconf,authordraft]{acmart}

\usepackage[ruled,vlined]{algorithm2e}
\usepackage{balance}

\AtBeginDocument{%
  \providecommand\BibTeX{{%
    \normalfont B\kern-0.5em{\scshape i\kern-0.25em b}\kern-0.8em\TeX}}}

\copyrightyear{2021}
\acmConference[]{}
\acmBooktitle{}
\acmPrice{15.00}
\acmISBN{978-1-4503-XXXX-X/18/06}
\setcopyright{none}
\renewcommand\footnotetextcopyrightpermission[1]{}
\begin{document}

\title{
Recovering the Unbiased Scene Graphs from the Biased Ones 
}

\author{Meng-Jiun Chiou}
\affiliation{%
  \institution{National University of Singapore}
%   \city{Singapore}
%   \country{Singapore}
}
\email{mengjiun.chiou@u.nus.edu}

\author{Henghui Ding}
\affiliation{%
  \institution{ByteDance AI Lab}
%   \city{Singapore}
%   \country{Singapore}
}
\email{henghui.ding@bytedance.com}

\author{Hanshu Yan}
\affiliation{%
  \institution{National University of Singapore}
%   \city{Singapore}
%   \country{Singapore}
}
\email{hanshu.yan@u.nus.edu}

\author{Changhu Wang}
\affiliation{%
 \institution{ByteDance AI Lab}
%   \city{Singapore}
%   \country{Singapore}
}
\email{changhu.wang@bytedance.com}

\author{Roger Zimmermann}
\affiliation{%
  \institution{National University of Singapore}
%   \city{Singapore}
%   \country{Singapore}
}
\email{rogerz@comp.nus.edu.sg}

\author{Jiashi Feng}
\affiliation{%
  \institution{National University of Singapore}
%   \city{Singapore}
%   \country{Singapore}
}
\email{elefjia@nus.edu.sg}

% \affiliation{%

% \affiliation{%

% \author{Anonymous Author(s)}

\renewcommand{\shortauthors}{}

\begin{abstract}
    Given input images, scene graph generation (SGG) aims to produce comprehensive, graphical representations describing visual relationships among salient objects.
    Recently, more efforts have been paid to the long tail problem in SGG; however, the imbalance in the fraction of missing labels of different classes, or \textit{reporting bias}, exacerbating the long tail is rarely considered and cannot be solved by the existing debiasing methods.
    In this paper we show that, due to the missing labels, SGG can be viewed as a ``Learning from Positive and Unlabeled data'' (PU learning) problem, where the reporting bias can be removed by recovering the unbiased probabilities from the biased ones by utilizing \textit{label frequencies}, \emph{i.e.,} the per-class fraction of labeled, positive examples in all the positive examples.
    To obtain accurate label frequency estimates, we propose Dynamic Label Frequency Estimation (DLFE) to take advantage of training-time data augmentation and average over multiple training iterations to introduce more valid examples.
    Extensive experiments show that DLFE is more effective in estimating label frequencies than a naive variant of the traditional estimate, and DLFE significantly alleviates the long tail and achieves state-of-the-art debiasing performance on the VG dataset.
    We also show qualitatively that SGG models with DLFE produce prominently more balanced and unbiased scene graphs.
    The source code will be publicly available\footnote{\url{https://github.com/coldmanck/recovering-unbiased-scene-graphs}}.
\end{abstract}

%   <concept>
%       <concept_id>10010147.10010178.10010224.10010225.10010227</concept_id>
%       <concept_desc>Computing methodologies~Scene understanding</concept_desc>
%       <concept_significance>500</concept_significance>
%       </concept>
%  </ccs2012>

% \ccsdesc[500]{Computing methodologies~Scene understanding}

% \keywords{Scene Graph Generation, Reporting Bias, Missing Labels, Long Tail}


\maketitle
\pagestyle{plain}

%%%%%%%%% BODY TEXT
\section{Introduction}
Scene graph generation (SGG) \cite{lu2016visual} aims to predict visual relationships in the form of (\textit{subject-predicate-object}) among salient objects in images.
SGG has been shown to be helpful for image captioning \cite{yao2018exploring,yang2019auto,li2019know}, visual question answering \cite{teney2017graph,shi2019explainable}, indoor scene understanding \cite{armeni20193d, chiou2020zero} and thus has been drawing increasing attention \cite{zellers2018neural,yang2018graph,herzig2018mapping,chen2019knowledge,chen2019scene,gu2019scene,chen2019soft,dornadula2019visual,zareian2020bridging,Khademi_Schulte_2020,tang2020unbiased,Lin_2020_CVPR,DBLP:conf/bmvc/WangPL20,yan2020pcpl,knyazev2020graphdensity,he2020learning,wang2020sketching,zareian2020learning,sharifzadeh2020classification,10.1145/3394171.3413566,9084259,10.1145/3394171.3413575,hung2020contextual,tian2020part,DBLP:journals/corr/abs-2001-04735,chiou2021visual,chiou2021st}.

\begin{figure}[t!]
\begin{center}
\includegraphics[width=\linewidth]{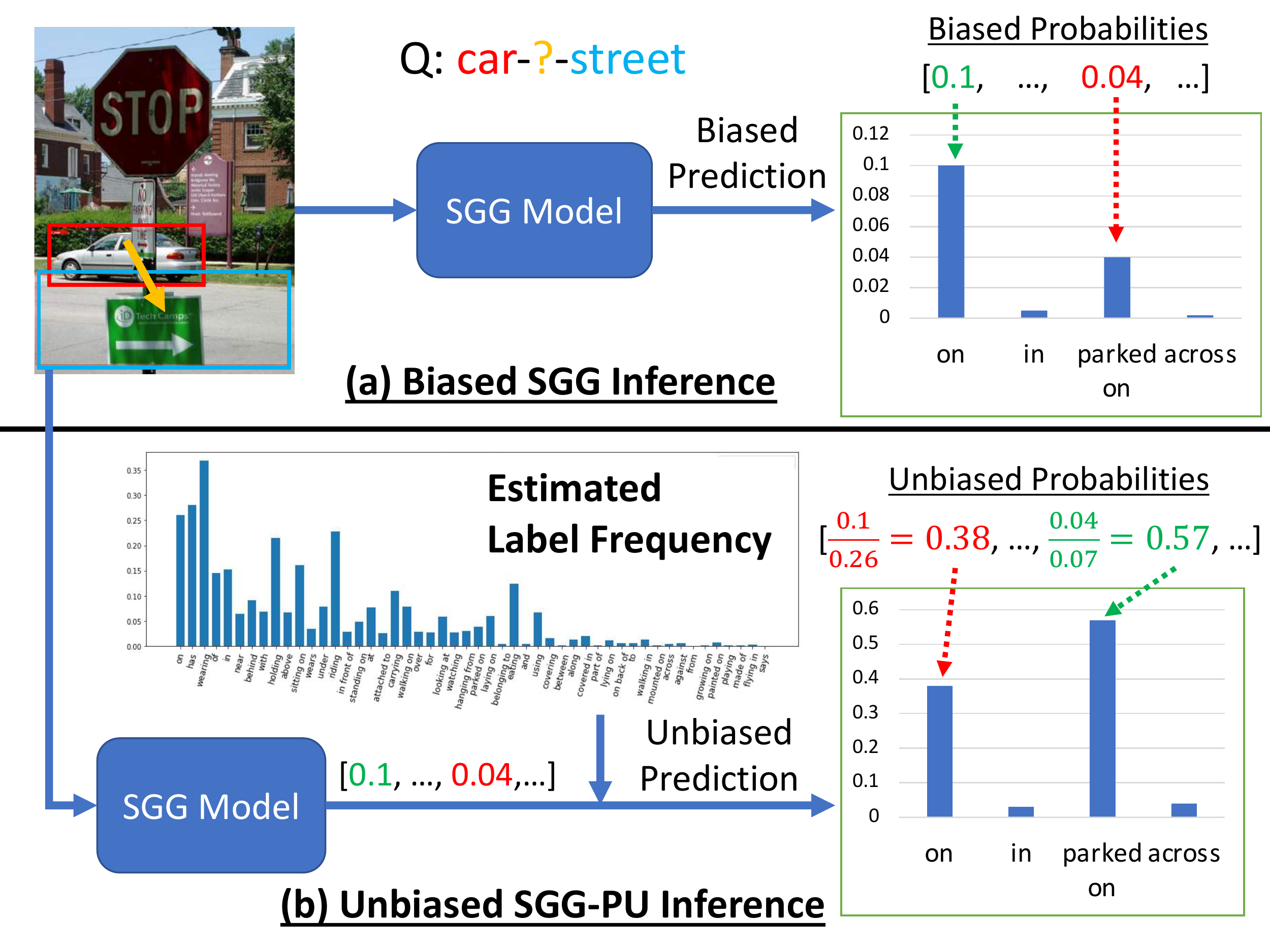}
\end{center}
\vspace{-1.5em}
  \caption{
  An illustrative comparison between the traditional, biased inference and the unbiased PU inference for SGG.
  (a) Traditionally SGG models are not trained in the PU setting and thus output biased probabilities in favor of conspicuous classes (\emph{e.g.}, {\fontfamily{qcr}\selectfont on}).
  (b) We remove the reporting bias from the biased probabilities by discounting the difference in the chance of being labeled, \emph{i.e.}, label frequency, so that inconspicuous classes (\emph{e.g.}, {\fontfamily{qcr}\selectfont parked on}) are properly predicted.
  }
\label{fig:motivation}
\vspace{-1.5em}
\end{figure}

The long tail problem is common and challenging in SGG \cite{tang2020unbiased}: since certain predicates (\emph{i.e.,} head classes) occur far more frequently than others (\emph{i.e.,} tail classes) in the most widely-used VG dataset \cite{krishna2017visual}, a model that trained with this unbalanced dataset would favor predicting heads against tails.
For instance, the number of training examples of {\fontfamily{qcr}\selectfont on} is $\sim$830$\times$ higher than that of {\fontfamily{qcr}\selectfont painted on} in the VG dataset, and (given ground truth objects) a classical SGG model MOTIFS \cite{zellers2018neural} achieves 74.3 Recall@20 for {\fontfamily{qcr}\selectfont on}, in sharp contrast to 0.0 for {\fontfamily{qcr}\selectfont painted on}.
However, the fact that the head classes are less descriptive than the tail classes makes the generated scene graphs coarse-grained and less informative, which is not ideal.

Most of the existing efforts in long-tailed SGG \cite{chen2019scene,tang2020unbiased,yan2020pcpl,DBLP:conf/bmvc/WangPL20,10.1145/3394171.3413575,he2020learning} deal with the skewed class distribution directly.
However, unlike common long-tailed classification tasks where the long tails are mostly caused by the unbalanced class prior distributions, the long tail of SGG with the VG dataset is significantly affected by the imbalance of missing labels, which remains unsolved.
The missing label problem arises as it is unrealistic to annotate the overwhelming number of possible visual relations (\emph{i.e.,} $KN(N-1)$ possibilities given $K$ predicate classes and $N$ objects in an image).
Training SGG models by treating all unlabeled pairs as negative examples (which is the default setting for most of the existing SGG works) introduces \textit{missing label bias} in predictions, \emph{i.e.,} predicted probabilities could be under-estimated.
What is worse, \textit{reporting bias} \cite{misra2016seeing,tang2020unbiased} which is prevalent in the VG dataset causes an imbalance in the missing labels of different predicates.
That is, the conspicuous classes (\emph{e.g.,} {\fontfamily{qcr}\selectfont on}, {\fontfamily{qcr}\selectfont in}) are more likely to be annotated than the inconspicuous ones (\emph{e.g.,} {\fontfamily{qcr}\selectfont parked on}, {\fontfamily{qcr}\selectfont covered in}).
Generally, conspicuous classes are more extensively annotated and have higher \textit{label frequencies}, \emph{i.e.,} the fraction of labeled, positive examples in all examples of individual classes.
The unbalanced label frequency distribution means that the predicted probability of an inconspicuous class could be under-estimated more than that of a conspicuous one, causing a long tail.
To produce meaningful scene graphs, the inconspicuous but informative predicates need to be properly predicted.
To the best of our knowledge, none of the existing SGG debiasing methods \cite{chen2019scene,tang2020unbiased,yan2020pcpl,DBLP:conf/bmvc/WangPL20,10.1145/3394171.3413575,he2020learning} effectively solve this reporting bias problem.

In this paper, we propose to tackle the reporting bias problem by removing the effect of unbalanced label frequency distribution.
That is, we aim to recover the unbiased version of per-class predicted probabilities such that they are independent of the per-class missing label bias.
To do this, we first show that learning a SGG model with the VG dataset can viewed as a \textit{Learning from Positive and Unlabeled data} (PU learning) \cite{denis2005learning,elkan2008learning,bekker2020learning} problem, where a target PU dataset contains only positive examples and unlabeled data.
For clarity, we define that a \textit{biased model} is trained on a PU dataset by treating the unlabaled data as negatives and outputs \textit{biased probabilities}, while an \textit{unbiased model} is trained on a fully-annotated dataset and outputs \textit{unbiased probabilities}.
Under the PU learning setting, the per-class unbiased probabilities are proportional to the biased ones with the per-class label frequencies as the proportionality constants \cite{elkan2008learning}.
Motivated by this fact, we propose to recover the unbiased visual relationship probabilities from the biased ones by dividing by the estimated per-class label frequencies so that the imbalance (\emph{i.e.,} reporting bias) can be offset.
Especially, the inconspicuous predicates with their probabilities being under-estimated more could then be predicted with higher confidences so that the scene graphs are more informative.
An illustrative comparison of the traditional, biased method and our unbiased one is shown in Fig. \ref{fig:motivation}. 

A traditional estimator of label frequencies is the per-class average of biased probabilities on a training/validation set predicted by a biased model \cite{elkan2008learning}.
While this estimator can work in the easier SGG settings where ground truth bounding boxes are given, \emph{i.e.}, PredCls and SGCls, it is found unable to provide estimates for some classes in the hardest SGG setting where no additional information other than images is provided, \emph{i.e.} SGDet.
The reason is that there are no \textit{valid} examples (\emph{i.e.}, predicted object pairs that match ground truth boxes and object labels) can be used for label frequency estimation.
For instance, by forwarding a trained MOTIFS \cite{zellers2018neural} model on VG training set, 9 out of 50 predicates do not have even a single valid example, making it impossible to estimate.
In this paper, we propose to take advantage of the training-time data augmentation such as random flipping to increase the number of valid examples.
That is, instead of performing post-training estimations, we propose Dynamic Label Frequency Estimation (DLFE) utilizing augmented training examples by maintaining a moving average of the per-batch biased probability during training.
The significant increase in the number of valid examples, especially in SGDet, enables accurate label frequency estimation for unbiased probability recovery.

Our contribution in this work is three-fold. 
First, we are among the first to tackle the long tail problem in SGG from the perspective of reporting bias, which we remove by recovering the per-class unbiased probability from the biased one with a PU based approach.
Second, to obtain accurate label frequency estimates for recovering unbiased probabilities in SGG, we propose DLFE which takes advantage of training-time data augmentation and averages over multiple training iterations to introduce more valid examples.
Third, we show that DLFE provides more reliable label frequency estimates than a naive variant of the traditional estimator, and we demonstrate that SGG models with DLFE effectively alleviates the long tail and achieve state-of-the-art debiasing performance with remarkably more informative scene graphs.
We will release the source code to facilitate research towards an unbiased SGG methodology.

\section{Related Work}
\subsection{Scene Graph Generation (SGG)}
SGG \cite{lu2016visual} aims to generate pairwise visual relationships in the form of (\textit{subject-predicate-object}) among salient objects, and there exists three training and evaluation settings \cite{xu2017scene,zellers2018neural}: (1) \textit{Predicate Classification} (PredCls) predicting relationships given ground truth bounding boxes and object labels, (2) \textit{Scene Graph Classification} (SGCls) predicting relationships and object labels given bounding boxes and (3) \textit{Scene Graph Detection} (SGDet) predicting relationships, object labels and bounding boxes with only input images.

Typically, SGG models consist of three main modules: proposal generation, object classification, and relationship prediction.
Generally, a pre-trained object detection model (\emph{e.g.,} \cite{ren2015faster}) is adopted for generating proposals.
For object classification, instead of using the predictions of object detection models directly, the generated proposals and their features are usually refined into \textit{object contexts} \cite{xu2017scene,yang2018graph,zellers2018neural,chen2019knowledge,tang2019learning} followed by decoding into object labels.
A common way to take object contexts into consideration is to run message passing algorithms (\emph{e.g.,} \cite{hochreiter1997long,tai-etal-2015-improved,li2015gated}) on a fully-connected \cite{xu2017scene,yang2018graph,chen2019knowledge}, chained \cite{zellers2018neural} or tree-structured \cite{tang2019learning} graph.
For relationship prediction, most approaches \cite{zellers2018neural,tang2020unbiased,chen2019knowledge} take in object contexts and bounding box features to compute \textit{relation contexts} with a similar graphical manner.
However, not until the recent works \cite{tang2019learning,chen2019knowledge} proposed the less biased \textit{mean recall} metrics did the research communities in SGG pay attention to the class imbalance problem \cite{gu2019scene,dornadula2019visual,Lin_2020_CVPR,tang2020unbiased,yan2020pcpl,DBLP:conf/bmvc/WangPL20}.
Tang et al. \cite{tang2020unbiased} borrow the counterfactual idea from causal inference to remove the context-specific bias.
Yan et al. \cite{yan2020pcpl} propose to perform re-weighting with class relatedness-aware weights.
Wang et al. \cite{DBLP:conf/bmvc/WangPL20} transfer the less-biased knowledge from the secondary learner to the main one with knowledge distillation.

Our proposed method can be viewed as a model-agnostic debiasing method \cite{tang2020unbiased,yan2020pcpl,DBLP:conf/bmvc/WangPL20}.
However, instead of focusing on 
class relatedness \cite{yan2020pcpl},
context co-occurrence bias \cite{tang2020unbiased} 
or missing label bias \cite{DBLP:conf/bmvc/WangPL20},
we tackle the underlying reporting bias \cite{misra2016seeing} by dealing with the unbalanced label frequency distribution.

\begin{figure*}[t!]
\begin{center}
\includegraphics[width=0.98\textwidth]{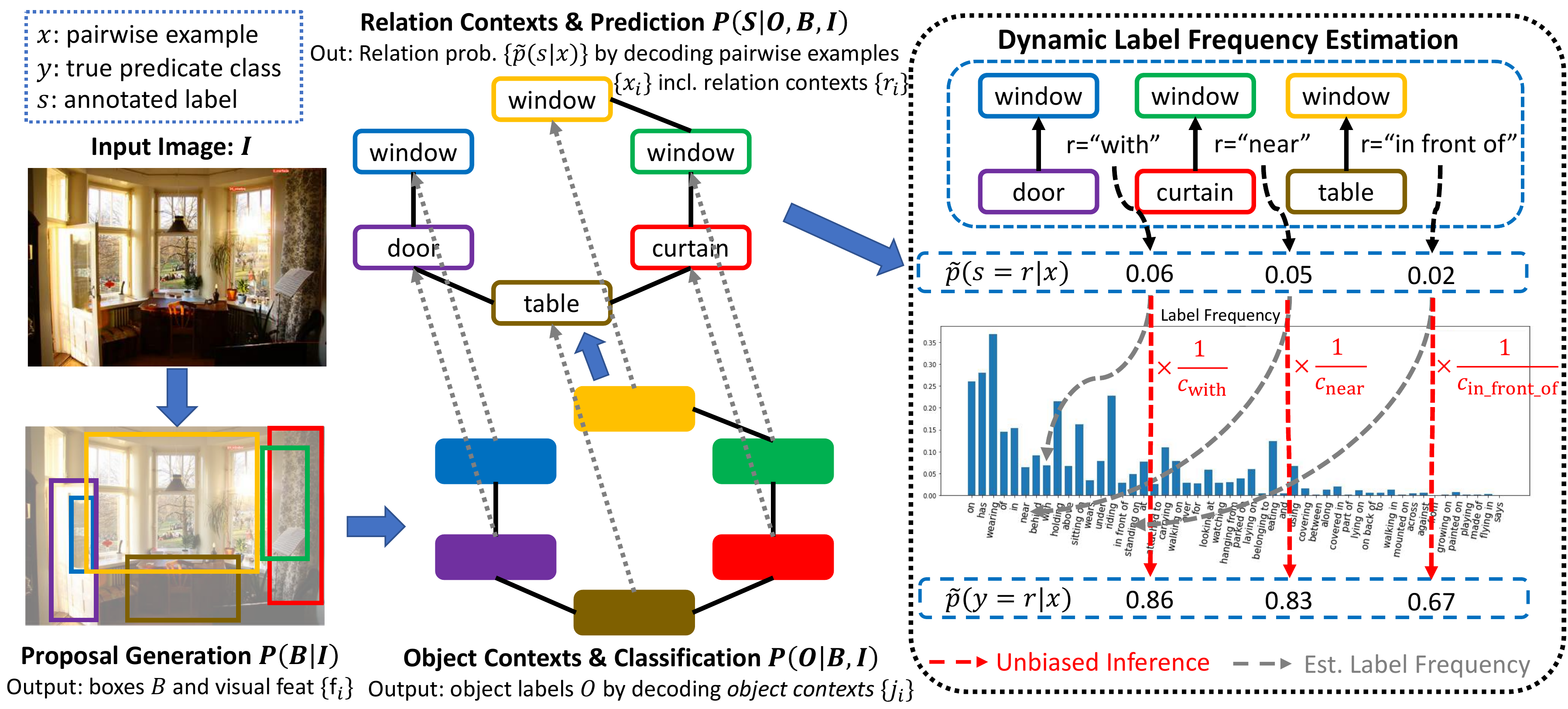}
\end{center}
\vspace{-1em}
  \caption{
  An illustration of training and inferencing a SGG model in a PU manner with Dynamic Label Frequency Estimation (DLFE).
  Given an input image, proposals and their features are extracted by an object detector. 
  Object classification is performed via message passing on a (\emph{e.g.,} chained \cite{zellers2018neural}) graph followed by \textit{object contexts} decoding.
  Object contexts together with bounding boxes and features are then fed into another graph to refine into \textit{relation contexts}, followed by decoding into the \textit{biased probabilities} $\tilde{p}(s|x)$.
  DLFE dynamically estimates the label frequencies $c$ with the moving averages of biased probabilities during training.
  Finally, the unbiased probability of class $r$ is recovered with $\tilde{p}(y=r|x)=\frac{1}{c_r} \tilde{p}(s=r|x)$ during inference.
    }
\label{fig:dlfe}
\vspace{-0.7em}
\end{figure*}

\subsection{Positive Unlabeled (PU) learning}
While the traditional classification setting aims to learn classifiers with both positive and negative data, \textit{Learning from Positive and Unlabeled data}, or \textit{Positive Unlabeled} (PU) learning, is a variant of the traditional setting where a PU dataset contains only positive and unlabeled examples \cite{denis2005learning,elkan2008learning,bekker2020learning}.
That is, an unlabeled example can either be truly a negative, or belongs to one or more classes.
Learning a biased classifier assuming all unlabeled examples are negative (which is the default setting for most of the existing SGG works) could introduce missing label bias, producing unbalanced predictions.
Common PU learning methods can be roughly divided into two categories \cite{bekker2020learning}: (a) training an unbiased model, and (b) inferencing a biased model in a PU manner.
We adopt the latter approach in this paper due to its convenience and favorable flexibility.

We note that while Chen et al. \cite{chen2019soft} also deal with SGG in the PU setting, they do not dive deep into the long tail problem in scene graphs as we do in this paper.
They propose a three-stage approach which generates pseudo-labels for the unlabeled examples with a biased trained model, followed by training a less biased model with the additional ``positive'' examples.
However, their approach is time and resource consuming since it requires re-generating pseudo labels if different SGG models are used.
Unlike \cite{chen2019soft}, our approach not only can be easily adapted for any SGG model with minimal modification, but is superior in terms of debiasing performance.

\section{Methodology}

Scene graph generation aims to generate a graph $G$ comprising bounding boxes $B$, object labels $O$, and visual relationships $S$, given an input image $I$.
The SGG task $P(G|I)$ is usually decomposed into three components for joint training \cite{zellers2018neural}:
\begin{equation}
    P(G|I)=P(B|I)P(O|B,I)P(S|O,B,I),
\end{equation}
where $P(B|I)$ denotes proposal generation, $P(O|B,I)$ means object classification and $P(S|O,B,I)$ is relationship prediction.
We propose to biasedly train a SGG model while we dynamically estimate the label frequencies during training.
The estimated label frequencies are then used to recover the unbiased probabilities during inference.

We describe our choice of proposal generation, object classification and relationship prediction in Section \ref{sec:model_components}.
We then explain how we recover the unbiased probabilities from the biased ones from a PU perspective in Section \ref{sec:preliminary_on_pu_learning}, followed by presenting our Dynamic Label Frequency Estimation (DLFE) in Section \ref{sec:dlfe}.
Figure \ref{fig:dlfe} shows an illustration of DLFE applied to SGG models like \cite{zellers2018neural,tang2019learning}.

\subsection{Model Components}
\label{sec:model_components}
\subsubsection{Proposal Generation $P(B|I)$}
Given an image $I$, we adopt a pre-trained object detector \cite{ren2015faster} to extract $N$ object proposals $B=\{b_i|i=1,...,N\}$, together with their visual representation $\{f_i|i=1,...,N\}$ and $N(N-1)$ union bounding box representations $\{f_{i,j}|i,j=1,...,N\}$ pooled from the output feature map.
The visual representations also come with predicted class probabilities: $\{p_i|i=1,...,N\}$ and $\{p_{i,j}|i,j=1,...,N\}$.

\subsubsection{Object Classification $P(O|B,I)$}
For object classification, a graphical representation is constructed which takes in object features $\{f_{i}\}$ and class probabilities $\{p_{i}\}$ and outputs \textit{object context} $\{j_i\}$ refined with message passing algorithms.
We experiment our methods with either chained-structured graphs \cite{zellers2018neural} with bi-directional LSTM \cite{hochreiter1997long}, or tree-structured graphs \cite{tang2019learning} with TreeLSTM \cite{tai-etal-2015-improved}.
The output object contexts are then fed into a linear layer followed with a Softmax layer to decode into predicted object labels $O=\{o_i|i=1,...,N\}$.

\subsubsection{Relationship Prediction $P(S|O,B,I)$}
Similar to that of object classification, another graphical representation of the same type is established to propagate contexts between features.
The module takes in both the object labels $O$ and the object contexts $\{j_i\}$ and outputs refined \textit{relation contexts} $\{r_i\}$.
For each object pair $\{(i, j)|i,j=1,...,N, i \neq j\}$, their relation contexts $(r_i, r_j)$, bounding boxes $(b_i, b_j)$, union bounding boxes $b_{ij}$ and features $f_{i,j}$ are gathered into a pairwise feature $x_{ij}$ for decoding into a probability vector over the $K$ classes with MLPs followed by a Softmax layer.

\begin{figure}[t!]
\begin{center}
\includegraphics[width=0.97\linewidth]{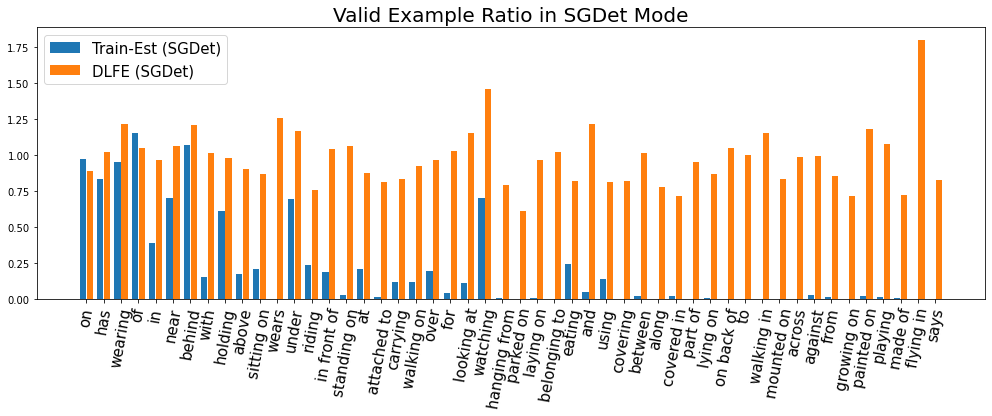}
\end{center}
\vspace{-1.7em}
  \caption{
  The per-class ratio of \textit{valid} examples in all examples of VG150 training set \cite{xu2017scene} (SGDet), by inferencing a trained MOTIFS (Train-Est) or dynamically inferencing a training MOTIFS with augmented data (DLFE).
  Numbers of DLFE are averaged over all epochs.
  All numbers can exceed 1 as a ground truth pair can match multiple proposal pairs.
}
\label{fig:valid_example_ratio_train-est_dlfe}
\vspace{-1.2em}
\end{figure}

\subsection{Recovering the Unbiased Scene Graphs}
\label{sec:preliminary_on_pu_learning}
Learning SGG from a dataset with missing labels can be viewed as a PU learning problem, which is different from the traditional classification in that (a) no negative examples are available, and (b) unlabeled examples can either be truly negatives or belong to any class.
Learning classifiers from a PU dataset by treating all unlabeled data as negatives could introduce strong \textit{missing label bias} \cite{elkan2008learning}, \emph{i.e.,} predicted probabilities could be under-estimated, and \textit{reporting bias} \cite{misra2016seeing}, \emph{i.e.,} predicted probability of an inconspicuous class could be under-estimated more than that of a conspicuous one.
We propose to avoid the both biases by recovering the unbiased probabilities, marginalizing the effect of uneven label frequencies.

Given $K$ predicate classes, we denote the visual relation examples taken in by the relationship prediction module of a SGG model by a set of tuple $(x, y, s)$, with $x$ an example (\emph{i.e.,} pairwise object features), $y \in \{0,...,K\}$ the true predicate class (0 means the background class) and $s \in \{0,...,K\}$ the relation label (0 means unannotated).
The class $y$ cannot be observed from the dataset: while we can derive $y=s$ if the example is labeled ($s \neq 0$), $y$ can be any number ranging from $0$ to $K$ for an unlabeled example ($s=0$).

For clarity, we now regard $x$, $y$ and $s$ as random variables.
For a target class $r \in \{1,...,K\}$, a biased SGG model is trained to predict the \textit{biased probability} $P(s=r|x)$, which can be derived as follows:
\begin{align}
    P(s=r|x) &= P(s=r, y=r|x) \quad \ \text{(by PU definition)} \\
             &= P(y=r|x)P(s=r|y=r,x), 
\end{align}
where $P(s=r|y=r,x)$ is the probability of example $x$ being selected to be labeled and is called \textit{propensity score} \cite{bekker2020learning}.
Dividing each side by $P(s=r|y=r,x)$ we obtain the \textit{unbiased probability} $P(y=r|x)$:
\begin{equation}
    P(y=r|x) = \frac{P(s=r|x)}{P(s=r|y=r,x)}. \label{eq:4}
\end{equation}
However, as it is unrealistic to obtain the propensity scores of each $x$, the existing works propose to \cite{elkan2008learning,chen2019soft} bypass the dependence on each $x$ by making the \textit{Selected Completely At Random} (SCAR) assumption \cite{bekker2020learning}: non-background examples are selected for labeling entirely at random regardless of $x$, \emph{i.e.,} the set of labeled examples is uniformly drawn from the set of positive examples.
This means that $P(s=r|y=r,x) = P(s=r|y=r)$ and Eqn \ref{eq:4} can be written as
\begin{equation}
    P(y=r|x) = \frac{P(s=r|x)}{P(s=r|y=r)},
\end{equation}
where $P(s=r|y=r)$ is the \textit{label frequency} of class $r$, or $c_r$, which is the fraction of labeled examples in all the examples of class $r$.
Notably, discounting the effect of per-class label frequencies in this way also removes the reporting bias.
Since label frequencies are usually not provided by annotators, an estimation is required.

\subsection{Dynamic Label Frequency Estimation}
\label{sec:dlfe}

One of the most common estimators of label frequencies, named Train-Est, is the per-class average of biased probabilities $\tilde{p}(s|x)$ predicted by a biased model \cite{elkan2008learning} (see full derivation in Appendix \ref{sec:appendix_derive_train_est}):
\begin{equation}
    c_r = P(s=r|y=r) \approx \frac{1}{N_r}\sum_{(x, y=r) \in D} \tilde{p}(s=r|x), \label{eq:6}
\end{equation}
where $D$ denotes a training or validation set and $N_r$ is the cardinality of $\{(x, y=r)\}$.
However, we find this way of estimation inconvenient and unsuitable for SGG.
To understand why, recall that PredCls, SGCls and SGDet are the three SGG training and evaluation settings, and note that re-estimation of label frequencies is required for each setting since the expected biased probabilities could vary depending on the task difficulty\footnote{Using label frequencies estimated in other mode is found to degrade the performance.}.
Firstly, the post-training estimation required before inferencing in each SGG setting is inconvenient and unfavorable.
Secondly, the absence of ground truth bounding boxes in SGDet mode results in lack of \textit{valid} examples for label frequency estimation.
For a proposal pair to be valid, its both objects must match ground truth boxes (with IoU $\geq 0.5$) and object labels simultaneously.
By using Train-Est with MOTIFS \cite{zellers2018neural}, as revealed in in Fig. \ref{fig:valid_example_ratio_train-est_dlfe} (the blue bars), 9 out of 50 predicates do not have even a valid example, \emph{i.e.,} $\{(x, y=r)\}$ in Eqn. \ref{eq:6} is empty, making it impossible to compute.
In addition, more valid examples are missing for inconspicuous classes: as the examples of those classes are concentrated in a much smaller number of images, not matching a bounding box could invalidate lots of examples.
A naive remedy is using a default value for those missing estimates; however, as we show in section \ref{sec:compare_dlfe_to_train_est} the performance is sub-optimal.

\begin{algorithm}[t]
\SetAlgoLined
\SetKwInOut{Input}{Input}\SetKwInOut{Output}{Output}
\Input{Training dataset $D^t$ and momentum $\alpha$}
\Output{Biased model $g(\cdot)$ and estimated label frequency $c$}
 
 \For{each mini batch $\mathcal{B} = \{(x_i,s_i)\} \in D^t$:}{
  Forward model to obtain the biased probabilities $g(x)$\;
  \tcp{in-batch average of biased probabilities}
  \For{each predicate class $r \in \{1,...,K\}$:}{
    $\mathcal{B}' \leftarrow \{(x_i,s_i)\in \mathcal{B}|s_i=r\}$\;
    $c_r' \leftarrow \frac{1}{|\mathcal{B}'|} \sum_{(x, s) \in \mathcal{B}'} g(s|x)$\;
    \tcp{Update the exponential moving average}
    $\tilde{c}_r\leftarrow\alpha\times c_r' + (1-\alpha)\times\tilde{c}_r$\;
  }
 }
%  \tcp{Save for inference use}
 $c \leftarrow \tilde{c}$\; 
 \caption{DLFE during training time}
 \label{alg:dlfe_train}
%   \vspace{-2em}
\end{algorithm}

To alleviate this problem, we propose to take advantage of the training-time data augmentation to get back more valid examples for tail classes.
Concretely, during training we augment input data by horizontal flipping with a probability of $0.5$, and meanwhile we estimate label frequencies with per-batch biased probabilities.
By doing this, the number of valid examples of tail classes could become more normal (and higher) than that of Train-Est, since averaging over augmented examples and multiple training iterations (with varying object label predictions) essentially introduces more samples, which in turn increases the number of valid examples.

Based on this idea, we propose Dynamic Label Frequency Estimation (DLFE) where the main steps are shown in Algorithm \ref{alg:dlfe_train}.
In detail, we maintain per-class moving averages of the biased probabilities (Eqn. \ref{eq:6}) throughout the training.
The estimated label frequencies $\tilde{c}$ are dynamically updated by the per-batch averages $c'$ with a momentum $\alpha$ so that the estimates that are more recent matter more: $\tilde{c} \leftarrow \alpha \times c' + (1 - \alpha) \times \tilde{c}$.
Note that for each mini-batch we update the estimated $\tilde{c}_r$ of class $r$ only if at least one valid example of $r$ presents in the current batch.
The estimated values gradually stabilize along with the converging model, and we save the final estimates $c \in \mathbb{R}^K$ (as a vector of length $K$).
During inference, the estimated label frequencies are utilized to recover the unbiased probability distribution $\tilde{p}(y|x)$ from the biased one $\tilde{p}(s|x)$ by
\begin{equation}
    \tilde{p}(y|x) = \frac{1}{c} \odot \tilde{p}(s|x),
\end{equation}
where $\odot$ denotes the Hadamard (element-wise) product.
The average per-epoch number of valid examples of this way is shown in Fig. \ref{fig:valid_example_ratio_train-est_dlfe} (the tangerine bars), where the inconspicuous classes get remarkably more ($4\times$ or more) valid examples.
This not only enables accurate estimations for all the classes but makes the estimations easier as no additional, post-training estimation is required.

\section{Experiments}

\subsection{Evaluation Settings}
\label{sec:eval_setting}
We follow the recent efforts in SGG \cite{zellers2018neural,chen2019knowledge} to experiment on a subset of the VG dataset \cite{krishna2017visual} named VG150 \cite{xu2017scene}, which contains $62,723$ images for training, $5,000$ for validation and $26,446$ for testing.
As discussed earlier, we train and evaluate in the three SGG settings: PredCls, SGCls and SGDet.
We evaluate models with, or without \textit{graph constraint}: whether only a single relation with the highest confidence is predicted for each object pair.
Non-graph constraint is denoted as ``ng''.
For evaluation, we adopt recall-based metrics which measures the fraction of ground truth visual relations appearing in top-$K$ confident predictions, where $K$ is 20, 50, or 100.
However, as the plain recall could be dominated by a biased model predicting mostly head classes, we follow \cite{chen2019knowledge,tang2019learning,yan2020pcpl,DBLP:conf/bmvc/WangPL20} to average over per-class recall and focus on the less biased per-class/mean recall (mR@$K$) and non-graph constraint per-class/mean recall (ng-mR@$K$)\footnote{While results in graph constraint recalls (R) and mean recalls (mR) are less reflective of how unbiased a SGG model is, we provide them for reference in Appendix \ref{sec:appendix_more_results_in_recalls}.}.
We note that the ng per-class/mean recall should be the fairest measure for debiasing methods since it 1) treats each class equally and 2) reflects the fact that more than one visual relations could exist for an object pair.
We follow the long-tailed recognition research \cite{liu2019large} to divide the distribution into three parts, including head (many-shot; top-15 frequent predicates), middle (medium-shot; mid-20) and tail (few-shot; last-15) and compute their ng-mRs.
Note that by DLFE in this section, we mean the dynamic label frequency estimation along with our unbiased scene graph recovery approach.

\subsection{Implementation Details}
\label{sec:impl_detail}
As DLFE is a model-agnostic strategy, we experiment with two popular SGG backbones: MOTIFS \cite{zellers2018neural} and VCTree \cite{tang2019learning}.
Following \cite{tang2020unbiased,DBLP:conf/bmvc/WangPL20}, we adopt a pre-trained and frozen Faster R-CNN \cite{ren2015faster} with ResNeXt-101-FPN \cite{lin2017feature} as the object detector, which achieves 28.14 mAP on VG's testing set \cite{tang2020unbiased}.
All the hyperparameters, including the momentum $\alpha=0.1$, are tuned with the validation set.
All models are trained using SGD optimizer with the initial learning rate of $0.01$ after the first $500$ iterations of warm-up.
Random flipping is applied to all the training examples.
% Validation is performed every $2,000$ iterations.
The learning rate is decayed, for a maximum of twice, by the factor of 10 once the validation performance plateaus twice consecutively.
Training can early stop when the maximum decay step (two) is reached before the maximum 50,000 iterations.
The final checkpoint is used for evaluation.
The batch size for all experiments is 48 (images).
For SGDet setting, we sample 80 proposals from each image and apply per-class NMS \cite{rosenfeld1971edge}.
Beside ground truth visual relations, we follow \cite{tang2020unbiased} to sample up to $1,024$ pairs with background-to-ground truth ratio being 3:1.

\subsection{Comparing DLFE to Train-Est}
\label{sec:compare_dlfe_to_train_est}

\begin{figure}[t!]
\begin{center}
\includegraphics[width=0.97\linewidth]{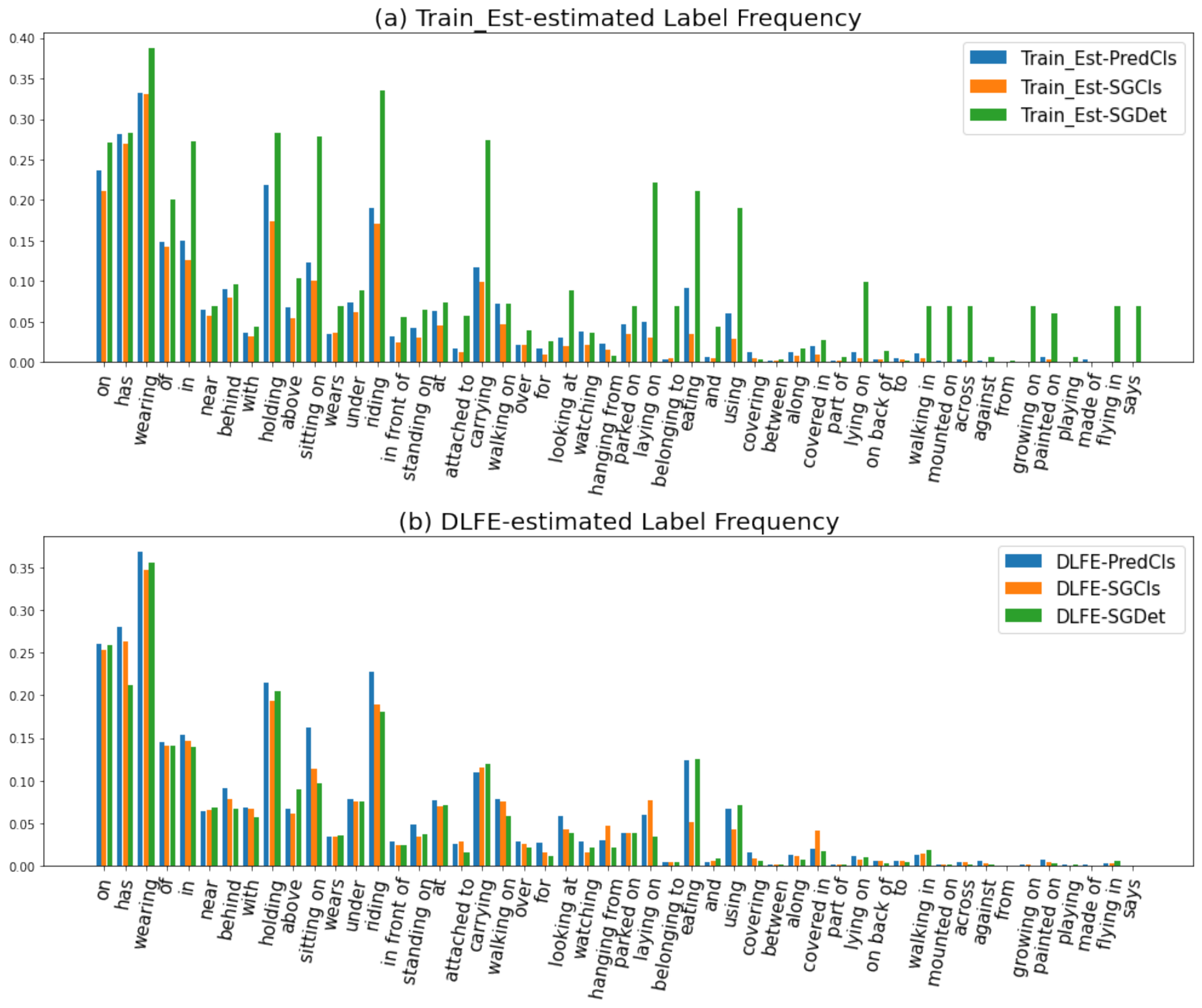}
\end{center}
\vspace{-1.8em}
  \caption{
  The label frequencies estimated by (a) Train-Est or (b) DLFE with MOTIFS \cite{zellers2018neural}.
  The classes with higher label frequency are more \textit{conspicuous} than those with a lower one.
  Predicates sorted by class frequency in descending order.
}
\label{fig:compare_label_freq}
\vspace{-0.7em}
\end{figure}

\begin{figure}[t!]
\begin{center}
\includegraphics[width=0.97\linewidth]{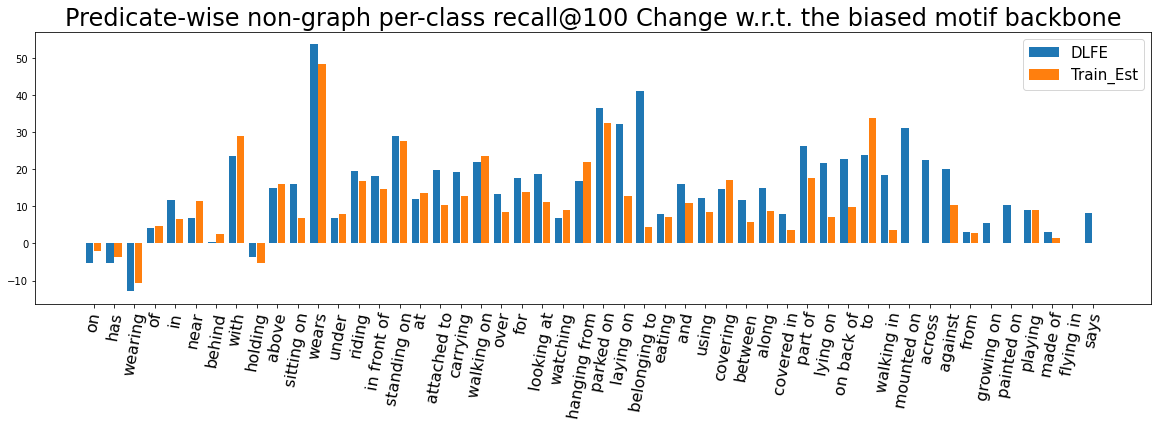}
\end{center}
\vspace{-1.8em}
  \caption{
  The absolute ng per-class R@100 changes when recovering MOTIFS's \cite{zellers2018neural} unbiased probabilities with the label frequencies estimated by Train-Est or DLFE, in SGDet mode.
}
\label{fig:ngr_compairson}
\vspace{-1.5em}
\end{figure}

\begin{table*}[htbp]
\centering
\scalebox{.9}{
\begin{tabular}{l|c c c|c c c|c c c}
\hline
\multicolumn{1}{c}{} & \multicolumn{3}{c}{Predicate Classification (PredCls)} & \multicolumn{3}{c}{Scene Graph Classification (SGCls)} & \multicolumn{3}{c}{Scene Graph Detection (SGDet)} \\
\hline
Model & ng-mR@20 & ng-mR@50 & ng-mR@100 & ng-mR@20 & ng-mR@50 & ng-mR@100 & ng-mR@20 & ng-mR@50 & ng-mR@100 \\ 
\hline 
KERN~\cite{chen2019knowledge} & - & 36.3 & 49.0 & - & 19.8 & 26.2 & - & 11.7 & 16.0 \\
% Schemata \cite{sharifzadeh2020classification} & - & 40.1 & 54.9 & - & 21.4 & 28.8 & - & - & - \\
GB-Net-$\beta$\textsuperscript{$\lozenge$} \cite{zareian2020bridging} & - & 44.5 & 58.7 & - & 25.6 & 32.1 & - & 11.7 & 16.6 \\
% GE-PCPL~\cite{yan2020pcpl} & - & 50.6 & 62.6 & - & 26.8 & 32.8 & - & 10.4 & 14.4 \\
\hline 
MOTIFS\textsuperscript{$\dagger$} \cite{zellers2018neural,DBLP:conf/bmvc/WangPL20} & 19.9 & 32.8 & 44.7 & 11.3 & 19.0 & 25.0 & 7.5 & 12.5 & 16.9 \\
% MOTIFS-Focal\textsuperscript{$\dagger$} \cite{lin2017focal,tang2020unbiased} & 10.9 & 13.9 & 15.0 & 6.3 & 7.7 & 8.3 & 3.9 & 5.3 & 6.6 \\
% MOTIFS-Resample\textsuperscript{$\dagger$} \cite{burnaev2015influence,tang2020unbiased} & 14.7 & 18.5 & 20.0 & 9.1 & 11.0 & 11.8 & 5.9 & 8.2 & 9.7 \\
MOTIFS-Reweight\textsuperscript{$\ddagger$} & 20.5 & 33.5 & 44.4 & 12.6 & 19.1 & 24.3 & 8.0 & 12.9 & 16.8 \\
MOTIFS-L2+uKD\textsuperscript{$\ddagger$} \cite{DBLP:conf/bmvc/WangPL20} & - & 36.9 & 50.9 & - & 22.7 & 30.1 & - & 14.0 & 19.5 \\
MOTIFS-L2+cKD\textsuperscript{$\ddagger$} \cite{DBLP:conf/bmvc/WangPL20} & - & 37.2 & 50.8 & - & 22.1 & 29.6 & - & 14.2 & 19.8 \\
MOTIFS-TDE\textsuperscript{$\dagger$} \cite{tang2020unbiased} & 18.7 & 29.0 & 38.2 & 10.7 & 16.1 & 21.1 & 7.4 & 11.2 & 14.9 \\
MOTIFS-PCPL\textsuperscript{$\dagger$} \cite{yan2020pcpl} & 25.6 & 38.5 & 49.3 & 13.1 & 19.9 & 25.6 & 9.8 & 14.8 & 19.6 \\
MOTIFS-STL\textsuperscript{$\dagger$} \cite{chen2019soft} & 15.7 & 29.4 & 43.2 & 10.3 & 18.4 & 27.2 & 6.4 & 10.6 & 15.0 \\
\textbf{MOTIFS-DLFE} & \textbf{30.0} & \textbf{45.8} & \textbf{57.7} & \textbf{17.6} & \textbf{25.6} & \textbf{32.0} & \textbf{11.7} & \textbf{18.1} & \textbf{23.0} \\
\hline
VCTree\textsuperscript{$\dagger$} \cite{tang2019learning,DBLP:conf/bmvc/WangPL20} & 21.4 & 35.6 & 47.8 & 14.3 & 23.3 & 31.4 & 7.5 & 12.5 & 16.7 \\
VCTree-Reweight\textsuperscript{$\ddagger$} & 20.6 & 32.5 & 41.6 & 14.1 & 21.3 & 27.8 & 8.0 & 12.1 & 15.9 \\
VCTree-L2+uKD\textsuperscript{$\ddagger$} \cite{DBLP:conf/bmvc/WangPL20} & - & 37.7 & 51.7 & - & 26.8 & 35.2 & - & 13.8 & 19.1 \\
VCTree-L2+cKD\textsuperscript{$\ddagger$} \cite{DBLP:conf/bmvc/WangPL20} & - & 38.4 & 52.4 & - & 26.8 & 35.8 & - & 13.9 & 19.0 \\
VCTree-TDE\textsuperscript{$\dagger$} \cite{tang2020unbiased} & 20.9 & 32.4 & 41.5 & 12.4 & 19.1 & 25.5 & 7.8 & 11.5 & 15.2 \\
VCTree-PCPL\textsuperscript{$\dagger$} \cite{yan2020pcpl} & 25.1 & 38.5 & 49.3 & 17.2 & 25.9 & 32.7 & 9.9 & 15.1 & 19.9 \\
VCTree-STL\textsuperscript{$\dagger$} \cite{chen2019soft} & 16.8 & 31.8 & 45.1 & 12.7 & 22.0 & 32.7 & 6.0 & 10.0 & 14.1 \\
\textbf{VCTree-DLFE} & \textbf{29.1} & \textbf{44.6} & \textbf{56.8} & \textbf{21.6} & \textbf{31.4} & \textbf{38.8} & \textbf{11.7} & \textbf{17.5} & \textbf{22.5} \\
\hline
\end{tabular}
}
\vspace{0.3em}
\caption{Performance comparison in ng-mR@$K$ on VG150 \cite{krishna2017visual,xu2017scene}.
Models in the first section are with VGG16 backbone \cite{simonyan2014very}.
$\dagger$ models implemented or reproduced ourselves with ResNeXt-101-FPN \cite{lin2017feature} backbone.
$\ddagger$ models also with the same ResNeXt-101-FPN backbone while their performance are reported by the respective papers.
\textsuperscript{$\lozenge$} model using external knowledge bases.
}
\vspace{-1.5em}
% \captionsetup[table]{skip=1pt}
\label{tab:sgg_ng_mr_result}
%\vspace{-1em}
\end{table*}

\begin{table*}[htbp]
\centering
\scalebox{.9}{
\begin{tabular}{l|c c c|c c c|c c c}
\hline
\multicolumn{1}{c}{} & \multicolumn{3}{c}{Predicate Classification (PredCls)} & \multicolumn{3}{c}{Scene Graph Classification (SGCls)} & \multicolumn{3}{c}{Scene Graph Detection (SGDet)} \\
\hline
Model  & mR@20 & mR@50 & mR@100 & mR@20 & mR@50 & mR@100 & mR@20 & mR@50 & mR@100 \\ 
\hline 
IMP+~\cite{xu2017scene, chen2019knowledge} & - & 9.8 & 10.5 & - & 5.8 & 6.0 & - & 3.8 & 4.8 \\
FREQ~\cite{zellers2018neural, tang2019learning} & 8.3 & 13.0 & 16.0 & 5.1 & 7.2 & 8.5 & 4.5 & 6.1 & 7.1 \\
MOTIFS~\cite{zellers2018neural, tang2019learning} & 10.8 & 14.0 & 15.3 & 6.3 & 7.7 & 8.2 & 4.2 & 5.7 & 6.6  \\
KERN~\cite{chen2019knowledge} & - & 17.7 & 19.2 & - & 9.4 & 10.0 & - & 6.4 & 7.3  \\
VCTree~\cite{tang2019learning} & 14.0 & 17.9 & 19.4 & 8.2 & 10.1 & 10.8 & 5.2 & 6.9 & 8.0 \\
% Schemata \cite{sharifzadeh2020classification} & - & 19.1 & 20.7 & - & 10.1 & 10.9 & - & - & - \\
GPS-Net \cite{Lin_2020_CVPR} & 17.4 & 21.3 & 22.8 & 10.0 & 11.8 & 12.6 & 6.9 & 8.7 & 9.8 \\
% GB-Net \cite{zareian2020bridging} & - & 19.3 & 20.9 & - & 9.6 & 10.2 & - & 6.1 & 7.3 \\
GB-Net-$\beta$\textsuperscript{$\lozenge$} \cite{zareian2020bridging} & - & 22.1 & 24.0 & - & 12.7 & 13.4 & - & 7.1 & 8.5 \\
\hline 
MOTIFS\textsuperscript{$\dagger$} \cite{zellers2018neural,tang2020unbiased} & 13.0 & 16.5 & 17.8 & 7.2 & 8.9 & 9.4 & 5.3 & 7.3 & 8.6 \\
MOTIFS-Focal\textsuperscript{$\ddagger$} \cite{lin2017focal,tang2020unbiased} & 10.9 & 13.9 & 15.0 & 6.3 & 7.7 & 8.3 & 3.9 & 5.3 & 6.6 \\
MOTIFS-Resample\textsuperscript{$\ddagger$} \cite{burnaev2015influence,tang2020unbiased} & 14.7 & 18.5 & 20.0 & 9.1 & 11.0 & 11.8 & 5.9 & 8.2 & 9.7 \\
MOTIFS-Reweight\textsuperscript{$\dagger$} & 14.3 & 17.3 & 18.6 & 9.5 & 11.2 & 11.7 & 6.7 & 9.2 & 10.9 \\
MOTIFS-L2+uKD\textsuperscript{$\ddagger$} \cite{DBLP:conf/bmvc/WangPL20} & 14.2 & 18.6 & 20.3 & 8.6 & 10.9 & 11.8 & 5.7 & 7.9 & 9.5 \\
MOTIFS-L2+cKD\textsuperscript{$\ddagger$} \cite{DBLP:conf/bmvc/WangPL20} & 14.4 & 18.5 & 20.2 & 8.7 & 10.7 & 11.4 & 5.8 & 8.1 & 9.6 \\
MOTIFS-TDE\textsuperscript{$\dagger$} \cite{tang2020unbiased} & 17.4 & 24.2 & 27.9 & 9.9 & 13.1 & 14.9 & 6.7 & 9.2 & 11.1 \\
MOTIFS-PCPL\textsuperscript{$\dagger$} \cite{yan2020pcpl} & 19.3 & 24.3 & 26.1 & 9.9 & 12.0 & 12.7 & 8.0 & 10.7 & 12.6 \\
MOTIFS-STL\textsuperscript{$\dagger$} \cite{chen2019soft} & 13.3 & 20.1 & 22.3 & 8.5 & 12.8 & 14.1 & 5.4 & 7.6 & 9.1 \\
\textbf{MOTIFS-DLFE} & \textbf{22.1} & \textbf{26.9} & \textbf{28.8} & \textbf{12.8} & \textbf{15.2} & \textbf{15.9} & \textbf{8.6} & \textbf{11.7} & \textbf{13.8} \\
\hline
VCTree\textsuperscript{$\dagger$} \cite{tang2019learning,tang2020unbiased} & 14.1 & 17.7 & 19.1 & 9.1 & 11.3 & 12.0 & 5.2 & 7.1 & 8.3 \\
VCTree-Reweight\textsuperscript{$\dagger$} & 16.3 & 19.4 & 20.4 & 10.6 & 12.5 & 13.1 & 6.6 & 8.7 & 10.1\\
VCTree-L2+uKD\textsuperscript{$\ddagger$} \cite{DBLP:conf/bmvc/WangPL20} & 14.2 & 18.2 & 19.9 & 9.9 & 12.4 & 13.4 & 5.7 & 7.7 & 9.2 \\ 
VCTree-L2+cKD\textsuperscript{$\ddagger$} \cite{DBLP:conf/bmvc/WangPL20} & 14.4 & 18.4 & 20.0 & 9.7 & 12.4 & 13.1 & 5.7 & 7.7 & 9.1 \\ 
VCTree-TDE\textsuperscript{$\dagger$} \cite{tang2020unbiased} & 19.2 & \textbf{26.2} & \textbf{29.6} & 11.2 & 15.2 & 17.5 & 6.8 & 9.5 & 11.4 \\
VCTree-PCPL\textsuperscript{$\dagger$} \cite{yan2020pcpl} & 18.7 & 22.8 & 24.5 & 12.7 & 15.2 & 16.1 & 8.1 & 10.8 & 12.6 \\
VCTree-STL\textsuperscript{$\dagger$} \cite{chen2019soft} & 14.3 & 21.4 & 23.5  & 10.5 & 14.6 & 16.6 & 5.1 & 7.1 & 8.4 \\
\textbf{VCTree-DLFE} & \textbf{20.8} & 25.3 & 27.1 & \textbf{15.8} & \textbf{18.9} & \textbf{20.0} & \textbf{8.6} & \textbf{11.8} & \textbf{13.8} \\
\hline
\end{tabular}
}
\vspace{0.1em}
\caption{Performance comparison of SGG models in graph-constraint mR@$K$ on VG150 \cite{krishna2017visual,xu2017scene} testing set. 
Models in the first section are with VGG16 while the others are with ResNeXt-101-FPN.
$\dagger$, $\ddagger$ and $\lozenge$ are with the same meanings as in Table \ref{tab:sgg_ng_mr_result}.
}
% \captionsetup[table]{skip=1pt}
\label{tab:sgg_mr_result}
\vspace{-1.7em}
\end{table*}

\begin{figure*}[t!]
\begin{center}
\includegraphics[width=0.95\textwidth]{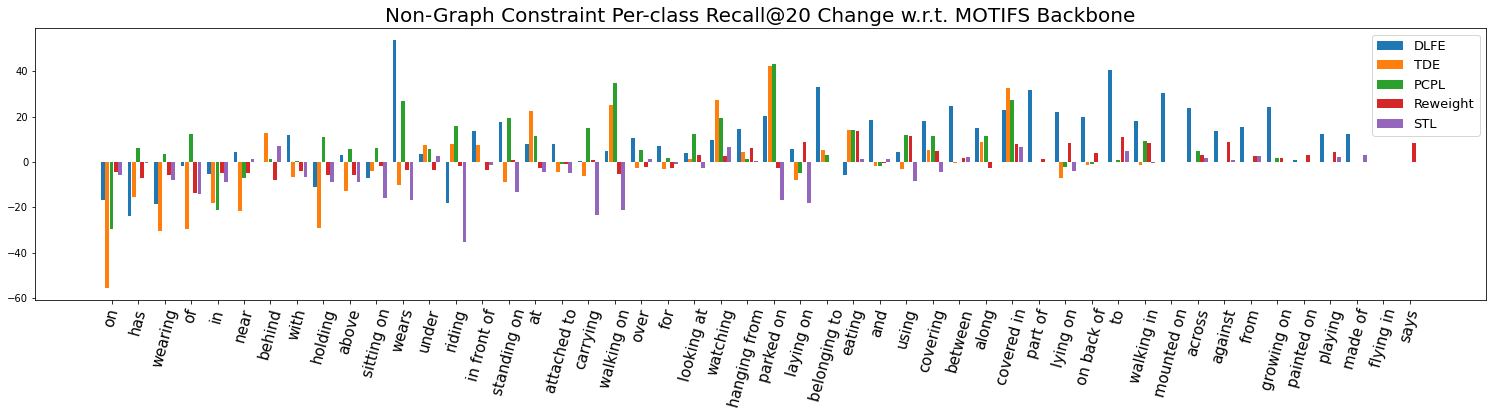}
\end{center}
\vspace{-1.2em}
  \caption{Non-graph constraint per-class Recall@20 (PredCls) change w.r.t. MOTIFS baseline.
%   Predicates are sorted by frequency from left to right.
  DLFE significantly improves the mid-to-tail recalls (where the other debiasing methods struggle) without compromising much head classes performance.}
\label{fig:recall_changes_to_smn}
\vspace{-1em}
\end{figure*}

We aim to answer the question: \textit{whether DLFE is more effective in estimating label frequency than Train-Est}, by comparing 1) the consistencies of the estimated label frequencies and 2) their debiasing performance.
As discussed earlier that label frequencies of the predicates lacking a valid example cannot be estimated by Train-Est, we thus naively assign the median of the other estimated label frequencies to those missing values.

A comparison of the estimated label frequencies is presented in Fig. \ref{fig:compare_label_freq}.
It is clear from (a) that, even for the classes that with at least one valid example, the Train-Est estimated values tend to be abnormally high in SGDet setting.
Note that while there might be differences in estimated values for different SGG settings, with a same backbone they should still be relatively similar.
In contrast, (b) shows DLFE-estimated values are more consistent across the three settings.
We also compare their debiasing performance in SGDet\footnote{Results in numbers and the other two SGG settings are provided in Appendix \ref{sec:appendix_more_results_train_est_dlfe}.} with the absolute ng per-class R@$100$ change in Fig. \ref{fig:ngr_compairson}.
Apparently, Train-Est barely improves the per-class recalls especially for tail classes that lacks enough valid examples, while DLFE achieves higher and more consistent improvement across all the predicates.

These results verify the claim that, apart from being more convenient (requiring no post-training estimation), DLFE is more effective than naive Train-Est for providing more reliable estimates.

\begin{table}[t!]
\centering
\scalebox{0.85}{
\begin{tabular}{l|c c|c c|c c}
\hline
\multicolumn{1}{c}{} & \multicolumn{2}{c}{Head Recalls} & \multicolumn{2}{c}{Middle Recalls} & \multicolumn{2}{c}{Tail Recalls} \\
\hline
Model & R@50 & R@100 & R@50 & R@100 & R@50 & R@100 \\ 
\hline 
MOTIFS\textsuperscript{$\dagger$} \cite{zellers2018neural,tang2020unbiased} & 65.9 & \textbf{78.6} & 30.0 & 45.4 & 3.3 & 9.7 \\
MOTIFS-Reweight\textsuperscript{$\dagger$} & 57.4 & 69.2 & 30.7 & 43.0 & 13.3 & 21.5 \\
MOTIFS-TDE\textsuperscript{$\dagger$} \cite{tang2020unbiased} & 48.3 & 60.8 & 34.9 & 46.1 & 1.8 & 5.3 \\
MOTIFS-PCPL\textsuperscript{$\dagger$} \cite{yan2020pcpl} & \textbf{66.5} & 77.6 & 41.8 & \textbf{55.2} & 6.0 & 13.2 \\
MOTIFS-STL\textsuperscript{$\dagger$} \cite{chen2019soft} & 56.4 & 70.0 & 24.1 & 39.8 & 9.6 & 21.2 \\
\textbf{MOTIFS-DLFE} & 61.9 & 72.4 & \textbf{42.8} & 54.2 & \textbf{31.8} & \textbf{44.6} \\
\hline
VCTree\textsuperscript{$\dagger$} \cite{tang2019learning,tang2020unbiased} & \textbf{67.5} & \textbf{79.8} & 34.3 & 50.0 & 5.5 & 12.7 \\
VCTree-Reweight\textsuperscript{$\dagger$} & 61.6 & 73.4 & 28.3 & 38.3 & 9.0 & 14.3 \\
VCTree-TDE\textsuperscript{$\dagger$} \cite{tang2020unbiased} & 54.8 & 67.5 & 37.9 & 49.1 & 2.5 & 5.4 \\
VCTree-PCPL\textsuperscript{$\dagger$} \cite{yan2020pcpl} & 64.5 & 75.9 & \textbf{42.6} & \textbf{54.2} & 6.9 & 16.1 \\
VCTree-STL\textsuperscript{$\dagger$} \cite{chen2019soft} & 57.6 & 71.1 & 26.1 & 41.8 & 13.8 & 23.5 \\
\textbf{VCTree-DLFE} & 57.5 & 68.3 & 36.0 & 48.2 & \textbf{26.5} & \textbf{38.1} \\
\hline
\end{tabular}
}
\vspace{0.1em}
\caption{Non-graph constraint head, middle and tail recalls (PredCls). $\dagger$ and $\ddagger$ are with the same meanings as in Table \ref{tab:sgg_ng_mr_result}.
DLFE improves the tail recalls by a large margin.
}
% \captionsetup[table]{skip=1pt}
\label{tab:head_mid_tail_predcls_short}
\vspace{-1.5em}
\end{table}

\begin{figure}[t!]
\begin{center}
\includegraphics[width=\linewidth]{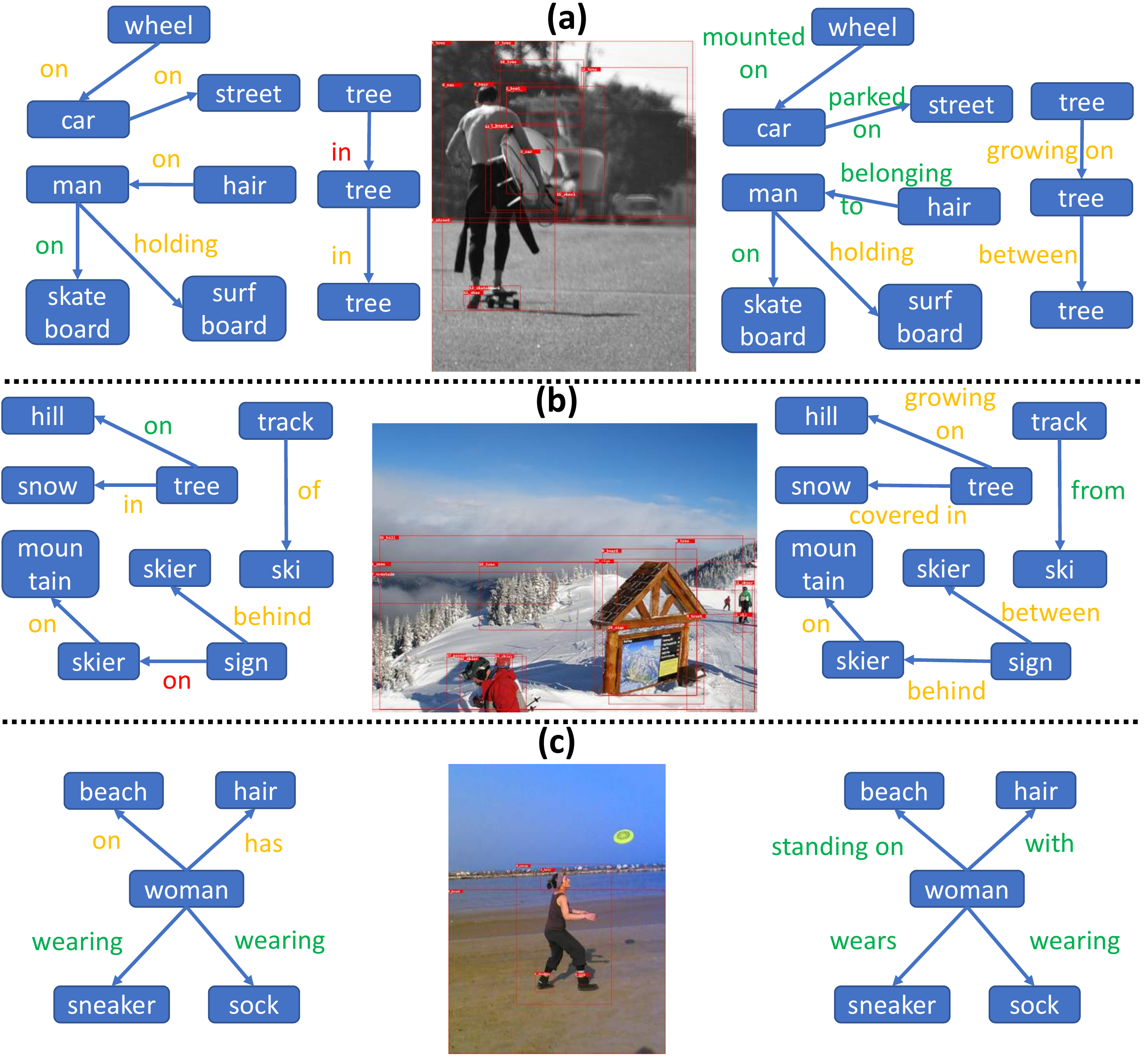}
\end{center}
\vspace{-1.2em}
  \caption{Scene graphs generated by MOTIFS (left) and MOTIFS-DLFE (right) in PredCls. 
  Only the top-1 prediction is shown for each object pair.
  A prediction can be \textcolor{Green}{correct} (matches GT), \textcolor{Red}{incorrect} (does not match GT and weird) or \textcolor{Dandelion}{acceptable} (does not match GT but still reasonable).
  }
\label{fig:sgg_vis}
\vspace{-1.5em}
\end{figure}

\subsection{Comparing to other Debiasing Methods}
\label{sec:compare_dlfe_to_sota}

While we list the results of different SGG backbone model for reference, we mainly compare our approach with the model-agnostic debiasing methods including \textit{Focal Loss} \cite{lin2017focal}, \textit{Resampling} \cite{burnaev2015influence}, \textit{Reweighting}, \textit{L2+\{u,c\}KD} \cite{DBLP:conf/bmvc/WangPL20}, \textit{TDE} \cite{tang2020unbiased}, \textit{PCPL} \cite{yan2020pcpl} and \textit{STL} \cite{chen2019soft}. 
L2+\{u,c\}KD is a two-learner knowledge distillation framework for reducing dataset biases.
TDE is an inference-time debiasing method which ensembles counterfactual thinking by removing context-specific bias.
PCPL learns the relatedness scores among predicates which are used as the weights in reweighting, and is the current state-of-the-art in terms of mR.
STL generates soft pseudo-labels for unlabeled data used for joint training.
We re-implement PCPL and STL since their backbone is not directly comparable, and Reweighting since there does not exist its reported performance for VCTree.
We report our reproduced results of TDE with the authors' codebase \cite{tang2020sggcode}.

\begin{figure*}[t!]
\begin{center}
\includegraphics[width=\textwidth]{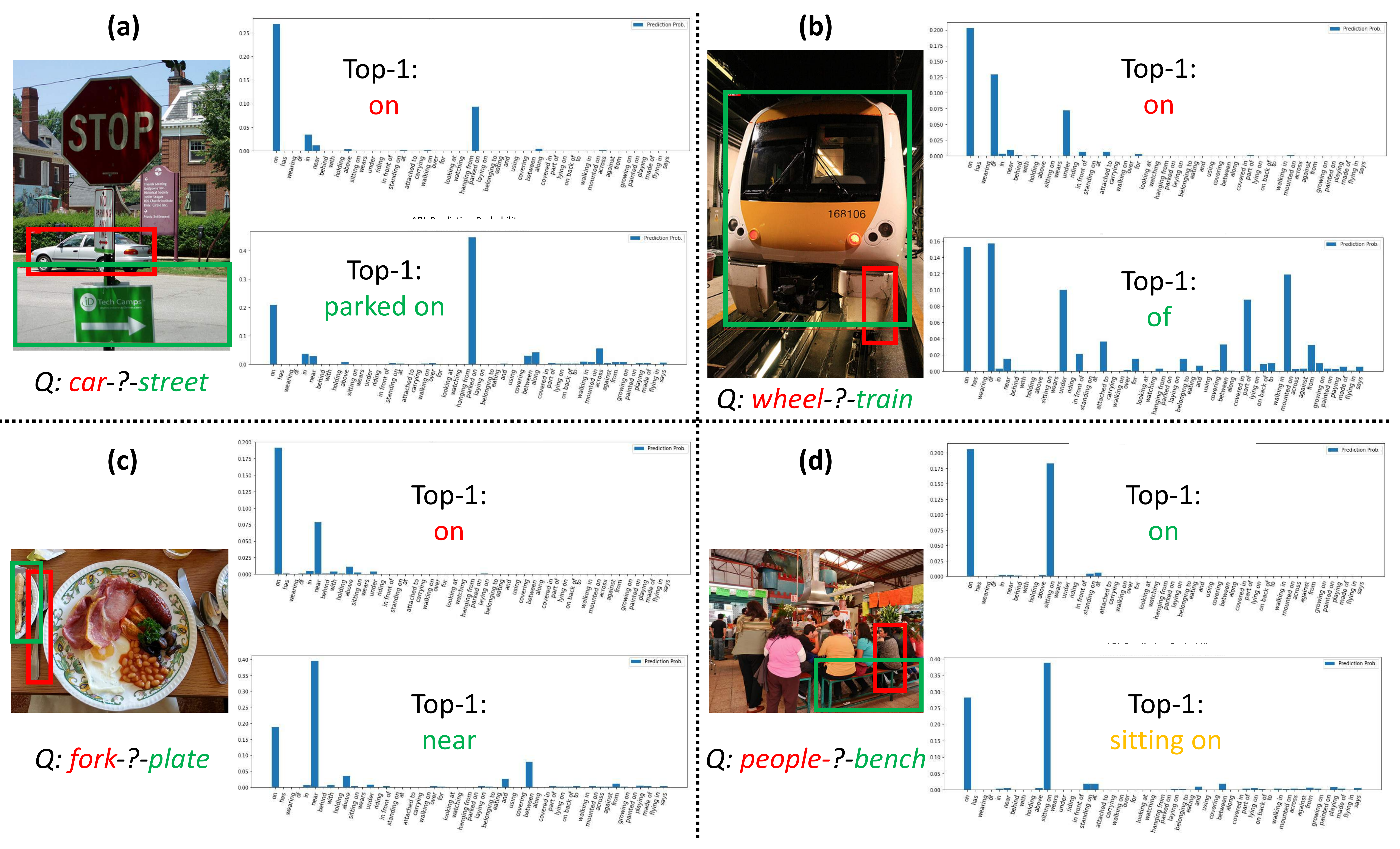}
\end{center}
\vspace{-1.8em}
  \caption{
  Probability distributions (normalized to sum 1) over the classes by MOTIFS (top of each example) and MOTIFS-DLFE (bottom). 
  The top-1 predictions can be \textcolor{Green}{correct} (GT), \textcolor{Red}{incorrect} (Non-GT and weird) or \textcolor{Dandelion}{acceptable} (Non-GT but reasonable).
  }
\label{fig:sgg_vis_dist}
\vspace{-1em}
\end{figure*}

The results in ng-mR@$K$ are presented in Table \ref{tab:sgg_ng_mr_result}, where DLFE significantly improves ng-mRs for both MOTIFS and VCTree and outperforms the existing debiasing methods.
Notably, TDE, which was proposed to alleviate the long tail by removing the context-specific bias, is shown to adversely affect the ability of predicting multiple relations for per object pair.
This shows that removing the reporting bias is more beneficial for debiasing SGG models.

While the graph-constraint mR metric does not reflect the fact that multiple relations could exist between an object pair, due to its popularity we still present the results in Table \ref{tab:sgg_mr_result}.
Debiasing MOTIFS with our proposed DLFE still significantly improves its mR, achieving state-of-the-art mR across all the three setting.
Large performance boost are also seen in VCTree with DLFE, and new SOTAs are attained for PredCls (mR@20), SGCls and SGDet.

To better understand how DLFE affects the performance of each class, we also present non-graph constraint per-class recall@20 changes compared to MOTIFS backbone-only (biased classifier), in Figure \ref{fig:recall_changes_to_smn}.
While all the debiasing methods increase the recall of the less frequent, middle-to-tail classes, only DLFE improves the tail (last-15) classes' performance significantly.
The other approach that also visibly improves the tail classes' performance is Reweighting; however, their relatively small improvements demonstrate that naively dealing with the unbalanced class frequencies is less effective than tackling the reporting bias.

We also present the head (many shot), middle (medium shot) and tail (few shot) non-graph constraint recalls in PredCls\footnote{The full results with graph-constraint, or SGCls/SGDet are available in Appendix \ref{sec:appendix_results_in_head_mid_tail}.} with the MOTIFS backbone in Table \ref{tab:head_mid_tail_predcls_short}.
Remarkably, DLFE outperforms the others by a significant margin regarding the tail recalls, \emph{e.g.,} Tail R@50 is 31.8 for DLFE, versus 1.8 for TDE, versus 6.0 for PCPL, versus 13.3 for Reweighting, showing that DLFE is especially capable of dealing with the long tail.

\subsection{Qualitative Results}
\label{sec:qualitative_results}

The scene graphs of three testing images are visualized in Figure \ref{fig:sgg_vis}, where the scene graphs on the left side are generated by MOTIFS and those on the right are by MOTIFS-DLFE.
(a) is an apparent example that, while {\fontfamily{qcr}\selectfont wheel-on-car}, {\fontfamily{qcr}\selectfont car-on-street}, {\fontfamily{qcr}\selectfont hair-on-man} predicted by MOTIFS are reasonable, {\fontfamily{qcr}\selectfont wheel-mounted on-car}, {\fontfamily{qcr}\selectfont car-parked on-street}, {\fontfamily{qcr}\selectfont hair-belonging to-man} predicted by MOTIFS-DLFE match the ground truth and are also more descriptive (while being inconspicuous).
Similarly, {\fontfamily{qcr}\selectfont tree-growing on-hill} in Example (b) and {\fontfamily{qcr}\selectfont woman-standing on-beach} in (c) are also correct and more descriptive; however, due to the missing label issue in the VG dataset, {\fontfamily{qcr}\selectfont tree-growing on-hill} can not be correctly recalled (shown as tangerine color).
In addition, there are some seemingly incorrect annotations such as {\fontfamily{qcr}\selectfont tree-standing on-tree} in Example (a), where the subject actually indicates a smaller, branch part of a tree.
For this object pair, MOTIFS-DLFE predicts {\fontfamily{qcr}\selectfont growing on} which, ironically, seems to be more reasonable than the ground truth label.

To understand how DLFE changes the probability distribution, we visualize the biased (MOTIFS) and unbiased (MOTIFS-DLFE) probabilities, given a subject-object pair, in Figure \ref{fig:sgg_vis_dist}.\footnote{More visualizations are available in Appendix \ref{sec:appendix_qualitative_results}.}
Prediction confidences are shown to be calibrated towards minor but expressive predicates like (a) {\fontfamily{qcr}\selectfont car-parked on-street}, (b) {\fontfamily{qcr}\selectfont wheel-of-train}, (d) {\fontfamily{qcr}\selectfont people-sitting on-bench} (while {\fontfamily{qcr}\selectfont sitting on} is not in the ground truth).
Notably in (c), {\fontfamily{qcr}\selectfont fork} is actually not {\fontfamily{qcr}\selectfont on} the plate but was mis-predicted by MOTIFS due to the strong bias (\emph{i.e.,} many {\fontfamily{qcr}\selectfont fork-in/on-plate} examples in VG dataset), while MOTIFS-DLFE correctly predicts {\fontfamily{qcr}\selectfont near}.
Moreover, (b) shows that the confidences of MOTIFS-DLFE for predicates other than the GT {\fontfamily{qcr}\selectfont of}, such as {\fontfamily{qcr}\selectfont mounted on} and {\fontfamily{qcr}\selectfont part of}, have increased remarkably, presumably because they are also reasonable choices.
This demonstrates the effectiveness of DLFE for balanced SGG.

\section{Conclusions}
In this paper, we deal with the long tail problem in SGG with the cause (unbalanced missing labels) instead of its superficial effect (long-tailed class distribution).
To ward off the reporting bias caused by the imbalance in missing labels, we view SGG as a PU learning problem and we remove the per-class missing label bias by recovering the unbiased probabilities from the biased ones.
To obtain reliable label frequencies for unbiased probability recovery, we take advantage of the data augmentation during training and perform Dynamic Label Frequency Estimation (DLFE) which maintains the moving averages of per-class biased probability and effectively introduces more valid samples, especially in SGDet training and evaluation mode.
Extensive quantitative and qualitative experiments demonstrate that DLFE is more effective in estimating label frequencies than a naive variant of the traditional estimator, and SGG models with DLFE achieve state-of-the-art debiasing performance on the VG dataset, producing well-balanced scene graphs.

\bibliographystyle{ACM-Reference-Format}
\balance
\bibliography{sample-sigconf}

\newpage

\nobalance

\phantomsection
\addcontentsline{toc}{chapter}{Appendices}
\section*{Appendices}
\setcounter{section}{0}
\renewcommand{\thesection}{\Alph{section}}
\renewcommand{\theHsection}{appendixsection.\Alph{section}}

\section{Deriving the Train-Est Estimator}
\label{sec:appendix_derive_train_est}
We show that why biased probabilities can be used as a label frequency estimator in this section. 
Denote an annotated example in a training or validation set with $(x, y, s)$, where $x$ is the pairwise example, $s$ is the relation label and $y$ is the true class.
Note $y=r$ as we are only considering the annotated ones.
Referring to \cite{elkan2008learning}, the biased probability $p(s|x)$ of class $r$ can be derived as follows:
\begin{align*}
    p(s=r|x) &= p(s=r|x, y=r)p(y=r|x)\\ 
             &\quad  + p(s=r|x, y\neq r)p(y\neq r|x) \\
             &= p(s=r|x, y=r)\times 1 + 0  \times 0\quad \text{(since y=r)}\\
             &= p(s=r|y=r), \quad \text{(the SCAR assumption)}
\end{align*}
where $p(s=r|y=r)$ is the label frequency of class $r$.
Thus, we can obtain a reasonable label frequency estimate via averaging the per-class biased probability with a training or validation set.

\section{More results of Train-Est and DLFE}
\label{sec:appendix_more_results_train_est_dlfe}
We present additional results in non-graph constraint mean recall (ng-mR@$K$) in Table \ref{tab:train_est_and_dlfe}.
Across both the MOTIFS \cite{zellers2018neural} and VCTree \cite{tang2019learning} backbone, our DLFE achieves significantly higher ng-mRs in both PredCls and SGDet setting and is on par with Train-Est in SGCls setting.

\begin{table*}[t!]
\centering
\def\arraystretch{1.1}%
\scalebox{0.9}{
\begin{tabular}{l|c c c|c c c|c c c}
\hline
\multicolumn{1}{c}{} & \multicolumn{3}{c}{Predicate Classification (PredCls)} & \multicolumn{3}{c}{Scene Graph Classification (SGCls)} & \multicolumn{3}{c}{Scene Graph Detection (SGDet)} \\
\hline
Model & ng-mR@20 & ng-mR@50 & ng-mR@100 & ng-mR@20 & ng-mR@50 & ng-mR@100 & ng-mR@20 & ng-mR@50 & ng-mR@100 \\ 
\hline 
MOTIFS \cite{zellers2018neural,tang2020unbiased} & 19.9 & 32.8 & 44.7 & 11.3 & 19.0 & 25.0 & 7.5 & 12.5 & 16.9 \\
MOTIFS-Train-Est \cite{elkan2008learning} & 24.4 & 38.9 & 50.5 & 17.1 & \textbf{26.1} & \textbf{32.8} & 8.9 & 14.1 & 18.9 \\
\textbf{MOTIFS-DLFE} & \textbf{30.0} & \textbf{45.8} & \textbf{57.7} & \textbf{17.6} & 25.6 & 32.0 & \textbf{11.7} & \textbf{18.1} & \textbf{23.0} \\
\hline
VCTree \cite{tang2019learning,tang2020unbiased} & 21.4 & 35.6 & 47.8 & 12.4 & 19.1 & 25.5 & 7.5 & 12.5 & 16.7 \\
VCTree-Train-Est \cite{elkan2008learning} & 25.0 & 39.1 & 52.4 & 21.0 & \textbf{32.2} & \textbf{39.4} & 8.1 & 13.0 & 17.1 \\
\textbf{VCTree-DLFE} & \textbf{29.1} & \textbf{44.6} & \textbf{56.8} & \textbf{21.6} & 31.4 & 38.8 & \textbf{11.7} & \textbf{17.5} & \textbf{22.5} \\
\hline
\end{tabular}
}
\vspace{0.1em}
\caption{Comparison of non-graph constraint mean recalls (ng-mR@$K$) between Train-Est and our DLFE, in PredCls, SGCls and SGDet. 
}
% \captionsetup[table]{skip=1pt}
\label{tab:train_est_and_dlfe}
\vspace{-1em}
\end{table*}

\section{Results in Head/Middle/Tail Recalls}
\label{sec:appendix_results_in_head_mid_tail}
\begin{figure*}[t!]
\begin{center}
\includegraphics[width=\textwidth]{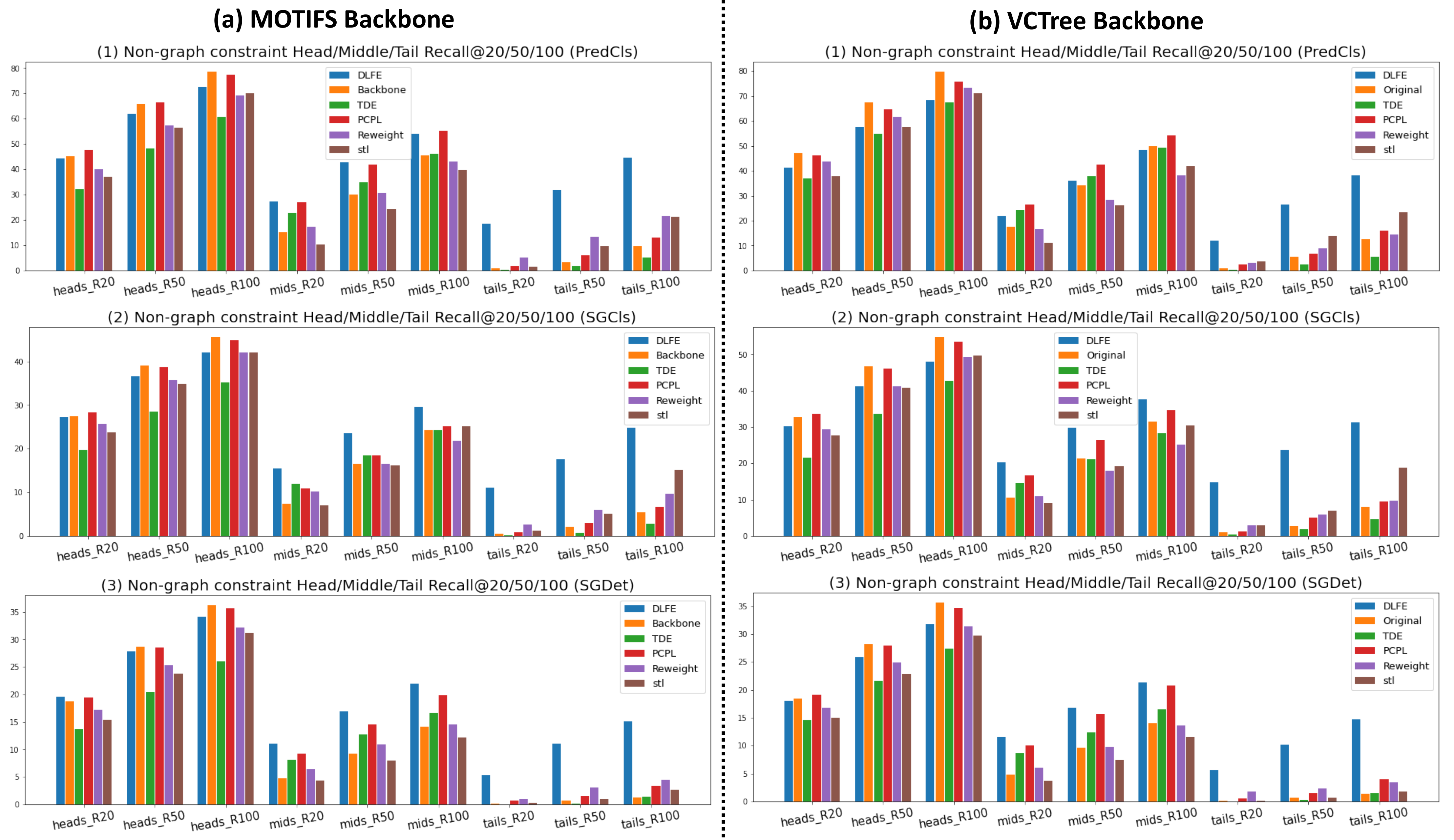}
\end{center}
\vspace{-1em}
  \caption{Bar plots of head (many shot), middle (medium shot) and tail (few shot) classes based on either MOTIFS (left) and VCTree (right) backbone, evaluated on VG150. From top to down is results in PredCls, SGCls and SGDet, respectively.}
\label{fig:head_mid_tail_vis_full}
% \vspace{-1em}
\end{figure*}

We compare our proposed DLFE with other debiasing methods, \emph{i.e.}, Reweighting, TDE \cite{tang2020unbiased}, PCPL \cite{yan2020pcpl} and STL \cite{chen2019soft}, on the recalls of different part of predicate distribution: (i) head (many-shot), (ii) middle (medium-shot) and (iii) tail (few-shot) recall.
The bar plots of head, middle and tail recalls (non-graph constraint) of MOTIFS and VCTree backbones are presented in Figure \ref{fig:head_mid_tail_vis_full}.
In all the three SGG tasks, both of MOTIFS-DLFE and VCTree-DLFE remarkably outperform other debiasing methods by a large margin regarding the tail recall, while being on par regarding the head and middle recalls.

The recalls with/without graph constraint in numbers for PredCls, SGCls and SGDet task is presented in Table \ref{tab:head_mid_tail_predcls}, Table \ref{tab:head_mid_tail_sgcls} and Table \ref{tab:head_mid_tail_sgdet}, respectively.
Again, DLFE improves middle and tail recalls more significantly, with the cost of head recall similar to that of other debiasing approaches.

\begin{table*}[htbp]
\centering
\def\arraystretch{1.1}%
\scalebox{0.93}{
\begin{tabular}{l|c c c|c c c|c c c}
\hline
\multicolumn{1}{c}{} & \multicolumn{9}{c}{Predicate Classification (PredCls)} \\
\multicolumn{1}{c}{} & \multicolumn{3}{c}{Head Recalls} & \multicolumn{3}{c}{Middle Recalls} & \multicolumn{3}{c}{Tail Recalls} \\
\hline
Model & R/ngR@20 & R/ngR@50 & R/ngR@100 & R/ngR@20 & R/ngR@50 & R/ngR@100 & R/ngR@20 & R/ngR@50 & R/ngR@100 \\ 
\hline 
MOTIFS\textsuperscript{$\dagger$} \cite{zellers2018neural,tang2020unbiased} & 35.5/45.2 & 42.9/65.9 & 45.4/\textbf{78.6} & 6.0/15.2 & 9.0/30.0 & 10.3/45.4 & 0.0/0.9 & 0.0/3.3 & 0.0/9.7 \\
MOTIFS-Reweight\textsuperscript{$\dagger$} & 32.4/40.0 & 38.5/57.4 & 40.6/69.2 & 10.1/17.3 & 12.6/30.7 & 13.9/43.0 & 1.9/5.3 & 2.4/13.3 & 2.9/21.5 \\
MOTIFS-TDE\textsuperscript{$\dagger$} \cite{tang2020unbiased} & 29.8/32.0 & 40.8/48.3 & 46.3/60.8 & 21.2/22.7 & 29.8/34.9 & 34.9/46.1 & 0.0/0.2 & 0.0/1.8 & 0.1/5.3 \\
MOTIFS-PCPL\textsuperscript{$\dagger$} \cite{yan2020pcpl} & \textbf{39.4}/\textbf{47.7} & \textbf{47.0}/\textbf{66.5} & \textbf{49.5}/77.6 & 19.5/27.0 & \textbf{25.3}/41.8 & \textbf{27.9}/\textbf{55.2} & 0.2/1.8 & 0.2/6.0 & 0.2/13.2 \\
MOTIFS-STL\textsuperscript{$\dagger$} \cite{chen2019soft} & 33.6/37.1 & 43.4/56.4 & 46.3/70.0 & 7.7/10.3 & 13.5/24.1 & 16.5/39.8 & 0.5/1.6 & 5.5/9.6 & 6.0/21.2 \\
\textbf{MOTIFS-DLFE} & 34.5/44.4 & 40.7/61.9 & 42.7/72.4 & \textbf{19.9}/\textbf{27.2} & \textbf{25.3}/\textbf{42.8} & \textbf{27.9}/54.2 & \textbf{11.2}/\textbf{18.6} & \textbf{15.1}/\textbf{31.8} & \textbf{15.7}/\textbf{44.6} \\
\hline
VCTree\textsuperscript{$\dagger$} \cite{tang2019learning,tang2020unbiased} & 36.6/\textbf{47.0} & 43.8/\textbf{67.5} & 46.2/\textbf{79.8} & 7.8/17.5 & 11.4/34.3 & 13.0/50.0 & 0.0/1.0 & 0.0/5.5 & 0.1/12.7 \\
VCTree-Reweight\textsuperscript{$\dagger$} & 36.6/43.6 & 43.1/61.6 & 45.4/73.4 & 12.2/16.6 & 14.6/28.3 & 15.2/38.3 & 1.3/2.9 & 2.3/9.0 & 2.3/14.3 \\
VCTree-TDE\textsuperscript{$\dagger$} \cite{tang2020unbiased} & 34.0/36.9 & 45.1/54.8 & \textbf{50.0}/67.5 & \textbf{22.5}/24.2 & \textbf{31.6}/37.9 & \textbf{36.2}/49.1 & 0.1/0.3 & 0.3/2.5 & 0.5/5.4 \\
VCTree-PCPL\textsuperscript{$\dagger$} \cite{yan2020pcpl} & \textbf{38.0}/46.3 & \textbf{45.4}/64.5 & 47.8/75.9 & 18.1/\textbf{26.3} & 23.0/\textbf{42.6} & 25.4/\textbf{54.2} & 0.1/2.4 & 0.1/6.9 & 0.1/16.1 \\
VCTree-STL\textsuperscript{$\dagger$} \cite{chen2019soft} & 34.0/37.8 & 43.8/57.6 & 46.5/71.1 & 8.4/10.9 & 14.3/26.1 & 17.1/41.8 & 2.4/3.6 & 8.4/13.8 & 9.1/23.5 \\
\textbf{VCTree-DLFE} & 30.1/41.1 & 37.3/57.5 & 39.6/68.3 & 14.4/21.9 & 19.0/36.0 & 20.8/48.2 & \textbf{6.0}/\textbf{12.0} & \textbf{8.1}/\textbf{26.5} & \textbf{9.2}/\textbf{38.1} \\
\hline
\end{tabular}
}
\vspace{0.1em}
\caption{Head, middle and tail (with/without graph constraint) recalls in the PredCls task on VG150. $\dagger$ and $\ddagger$ are with the same meaning as in Table 1 of the main paper.
}
% \captionsetup[table]{skip=1pt}
\label{tab:head_mid_tail_predcls}
\vspace{1em}
\end{table*}

\begin{table*}[htbp]
\centering
\def\arraystretch{1.1}%
\scalebox{0.93}{
\begin{tabular}{l|c c c|c c c|c c c}
\hline
\multicolumn{1}{c}{} & \multicolumn{9}{c}{Scene Graph Classification (SGCls)} \\
\multicolumn{1}{c}{} & \multicolumn{3}{c}{Head Recalls} & \multicolumn{3}{c}{Middle Recalls} & \multicolumn{3}{c}{Tail Recalls} \\
\hline
Model & R/ngR@20 & R/ngR@50 & R/ngR@100 & R/ngR@20 & R/ngR@50 & R/ngR@100 & R/ngR@20 & R/ngR@50 & R/ngR@100 \\
\hline 
MOTIFS\textsuperscript{$\dagger$} \cite{zellers2018neural,tang2020unbiased} & 21.3/\textbf{27.4} & 25.1/\textbf{39.1} & 26.3/\textbf{45.6} & 2.1/7.3 & 3.5/16.6 & 3.9/24.3 & 0.0/0.5 & 0.0/2.1 & 0.0/5.4 \\
MOTIFS-Reweight\textsuperscript{$\dagger$} & 21.6/25.7 & 25.1/35.7 & 26.3/42.1 & 6.8/10.2 & 8.0/16.5 & 8.3/21.9 & 0.9/2.6 & 1.6/6.0 & 1.7/9.7 \\
MOTIFS-TDE\textsuperscript{$\dagger$} \cite{tang2020unbiased} & 18.0/19.6 & 23.5/28.4 & 26.1/35.2 & 11.2/12.0 & \textbf{15.2}/18.5 & \textbf{17.8}/24.3 & 0.0/0.2 & 0.0/0.6 & 0.0/2.7 \\
MOTIFS-PCPL\textsuperscript{$\dagger$} \cite{yan2020pcpl} & \textbf{23.0}/28.2 & \textbf{27.1}/38.7 & \textbf{28.3}/44.9 & 7.3/10.9 & 9.5/18.5 & 10.3/25.2 & 0.1/0.9 & 0.1/3.1 & 0.2/6.6 \\
MOTIFS-STL\textsuperscript{$\dagger$} \cite{chen2019soft} & 21.3/23.8 & 26.4/34.8 & 27.8/42.1 & 5.2/7.0 & 8.8/16.2 & 10.2/25.2 & 0.2/1.2 & 4.4/5.2 & 5.7/15.1 \\
\textbf{MOTIFS-DLFE} & 21.2/27.2 & 24.6/36.6 & 25.6/42.0 & \textbf{11.3}/\textbf{15.4} & 14.2/\textbf{23.5} & 15.1/\textbf{29.5} & \textbf{7.1}/\textbf{11.1} & \textbf{8.3}/\textbf{17.7} & \textbf{8.4}/\textbf{24.8} \\
\hline
VCTree\textsuperscript{$\dagger$} \cite{tang2019learning,tang2020unbiased} & 25.3/32.8 & 29.9/\textbf{46.8} & 31.3/\textbf{54.7} & 3.8/10.5 & 5.8/21.2 & 6.4/31.5 & 0.0/1.0 & 0.0/2.6 & 0.0/8.0 \\
VCTree-Reweight\textsuperscript{$\dagger$} & 24.2/29.5 & 28.4/41.2 & 29.7/49.3 & 7.1/10.9 & 8.5/18.0 & 9.0/25.2 & 1.7/2.9 & 1.9/5.9 & 1.9/9.7 \\
VCTree-TDE\textsuperscript{$\dagger$} \cite{tang2020unbiased} & 19.1/21.6 & 26.2/33.6 & 30.0/42.7 & 13.7/14.5 & 18.2/21.2 & \textbf{21.2}/28.2 & 0.0/0.5 & 0.1/1.9 & 0.1/4.6 \\
VCTree-PCPL\textsuperscript{$\dagger$} \cite{yan2020pcpl} & \textbf{27.2}/\textbf{33.6} & \textbf{32.0}/46.2 & \textbf{33.5}/53.4 & 11.3/16.8 & 13.9/26.3 & 15.2/34.7 & 0.1/1.3 & 0.1/5.1 & 0.1/9.4 \\
VCTree-STL\textsuperscript{$\dagger$} \cite{chen2019soft} & 24.8/27.6 & 30.7/40.8 & 32.3/49.7 & 6.6/9.0 & 10.3/19.1 & 12.2/30.5 & 1.6/2.8 & 4.1/6.9 & 6.6/18.7 \\
\textbf{VCTree-DLFE} & 22.8/30.2 & 26.9/41.2 & 28.3/48.0 & \textbf{15.4}/\textbf{20.4} & \textbf{18.5}/\textbf{29.9} & 19.8/\textbf{37.6} & \textbf{9.3}/\textbf{14.7} & \textbf{11.3}/\textbf{23.8} & \textbf{12.0}/\textbf{31.3} \\
\hline
\end{tabular}
}
\vspace{0.1em}
\caption{Head, middle and tail (with/without graph constraint) recalls in the SGCls task on VG150. $\dagger$ and $\ddagger$ are with the same meaning as in Table 1 of the main paper.
}
% \captionsetup[table]{skip=1pt}
\label{tab:head_mid_tail_sgcls}
% \vspace{-1em}
\end{table*}

\begin{table*}[htbp]
\centering
\def\arraystretch{1.1}%
\scalebox{0.93}{
\begin{tabular}{l|c c c|c c c|c c c}
\hline
\multicolumn{1}{c}{} & \multicolumn{9}{c}{Scene Graph Detection (SGDet)} \\
\multicolumn{1}{c}{} & \multicolumn{3}{c}{Head Recalls} & \multicolumn{3}{c}{Middle Recalls} & \multicolumn{3}{c}{Tail Recalls} \\
\hline
Model & R/ngR@20 & R/ngR@50 & R/ngR@100 & R/ngR@20 & R/ngR@50 & R/ngR@100 & R/ngR@20 & R/ngR@50 & R/ngR@100 \\
\hline 
MOTIFS\textsuperscript{$\dagger$} \cite{zellers2018neural,tang2020unbiased} & 15.4/18.8 & 20.7/\textbf{28.7} & 24.1/\textbf{36.2} & 1.6/4.7 & 2.8/9.2 & 3.3/14.2 & 0.0/0.1 & 0.0/0.7 & 0.0/1.3 \\
MOTIFS-Reweight\textsuperscript{$\dagger$} & 15.3/17.2 & 20.4/25.2 & 23.9/32.2 & 4.8/6.5 & 6.4/10.9 & 7.9/14.6 & 0.5/0.9 & 1.8/3.1 & 1.9/4.5 \\
MOTIFS-TDE\textsuperscript{$\dagger$} \cite{tang2020unbiased} & 12.8/13.7 & 17.4/20.4 & 20.9/26.1 & 7.1/8.2 & 9.9/12.7 & 12.1/16.6 & 0.0/0.0 & 0.0/0.1 & 0.0/1.4 \\
MOTIFS-PCPL\textsuperscript{$\dagger$} \cite{yan2020pcpl} & \textbf{17.1}/19.5 & \textbf{22.5}/28.5 & \textbf{26.0}/35.7 & 7.1/9.3 & 9.8/14.5 & 12.0/19.8 & 0.1/0.7 & 0.1/1.6 & 0.1/3.3\\
MOTIFS-STL\textsuperscript{$\dagger$} \cite{chen2019soft} & 14.1/15.4 & 19.2/23.8 & 22.9/31.2 & 3.0/4.3 & 4.3/7.9 & 5.4/12.2 & 0.0/0.3 & 0.2/1.0 & 0.4/2.6 \\
\textbf{MOTIFS-DLFE} & 15.1/\textbf{19.6} & 20.0/27.9 & 23.4/34.1 & \textbf{8.0}/\textbf{11.0} & \textbf{10.9}/\textbf{16.9} & \textbf{13.3}/\textbf{21.9} & \textbf{3.5}/\textbf{5.3} & \textbf{5.3}/\textbf{11.1} & \textbf{6.7}/\textbf{15.1} \\
\hline
VCTree\textsuperscript{$\dagger$} \cite{tang2019learning,tang2020unbiased} & 15.1/18.5 & 20.1/\textbf{28.2} & 23.3/\textbf{35.6} & 1.8/4.8 & 2.6/9.6 & 3.3/14.0 & 0.0/0.1 & 0.0/0.6 & 0.0/1.3 \\
VCTree-Reweight\textsuperscript{$\dagger$} & 14.7/16.8 & 19.6/24.9 & 22.7/31.3 & 4.6/6.0 & 6.1/9.8 & 7.2/13.7 & 1.1/1.8 & 1.2/2.3 & 1.3/3.4 \\
VCTree-TDE\textsuperscript{$\dagger$} \cite{tang2020unbiased} & 12.9/14.5 & 17.8/21.7 & 21.4/27.3 & 7.3/8.6 & 10.3/12.4 & 12.5/16.5 & 0.0/0.1 & 0.0/0.2 & 0.0/1.4 \\
VCTree-PCPL\textsuperscript{$\dagger$} \cite{yan2020pcpl} & \textbf{16.8}/\textbf{19.1} & \textbf{22.0}/27.9 & \textbf{25.4}/34.7 & 7.6/10.0 & 10.3/15.7 & 12.4/20.7 & 0.1/0.5 & 0.1/1.5 & 0.1/4.0 \\
VCTree-STL\textsuperscript{$\dagger$} \cite{chen2019soft} & 13.6/15.0 & 18.5/22.8 & 21.8/29.7 & 2.4/3.7 & 3.8/7.4 & 4.7/11.5 & 0.1/0.2 & 0.1/0.7 & 0.1/1.8 \\
\textbf{VCTree-DLFE} & 13.6/18.0 & 18.2/25.9 & 21.2/31.9 & \textbf{8.3}/\textbf{11.5} & \textbf{11.3}/\textbf{16.8} & \textbf{13.1}/\textbf{21.3} & \textbf{3.9}/\textbf{5.6} & \textbf{6.0}/\textbf{10.2} & \textbf{7.2}/\textbf{14.7} \\
\hline
\end{tabular}
}
\vspace{0.1em}
\caption{Head, middle and tail (with/without graph constraint) recalls in the SGDet task on VG150. $\dagger$ and $\ddagger$ are with the same meaning as in Table 1 of the main paper.
}
% \captionsetup[table]{skip=1pt}
\label{tab:head_mid_tail_sgdet}
% \vspace{-1em}
\end{table*}

\section{More Results in Recalls}
\label{sec:appendix_more_results_in_recalls}
We present a comprehensive comparison of recently published (2020-now) SGG models/debiasing methods\footnote{Note that we do not compare with \cite{knyazev2020graphdensity} as 1) their reported numbers are rather selective and incomplete and 2) their method were not compared with other debiasing methods (\emph{e.g.,} \cite{tang2020unbiased}) fairly, \emph{i.e.,} with the same backbone.} in graph-constraint recalls and mean recalls, in PredCls mode in Table \ref{tab:sgg_result_predcls}, in SGCls mode in Table \ref{tab:sgg_result_sgcls} and in SGDet mode in Table \ref{tab:sgg_result_sgdet}.
% Only recalls with graph-constraint are presented since most of the existing methods
We note that the plain recall (R@$K$) is biased as it favors the head classes and does not reflect that multiple relations could exist in an object pair. 
While there are some performance drops in more conventional recalls (R@$K$) for the debiasing methods, it is because the predicates are being classified into the more descriptive ones (which do not have been annotated as ground truth).
Mean recall (mR@$K$) is less biased than the plain recall as it treats all classes equally; however, it still does not consider the multi-relation issue.
Remarkably, it is clear from all the three tables that our DLFE still achieves state-of-the-art mR@$K$ comparing the other debiasing methods with the same backbone, and either MOTIFS-DLFE or VCTree-DLFE attains the highest mR scores across all the models and backbones.

\begin{table*}[htbp]
\centering
\scalebox{1.0}{
\begin{tabular}{l|c c c c c c}
\hline
\multicolumn{1}{c}{} & \multicolumn{6}{c}{Predicate Classification} \\
\hline
Model & R@20 & R@50 & R@100 & mR@20 & mR@50 & mR@100\\ 
\hline 
IMP+~\cite{xu2017scene, chen2019knowledge} & 52.7 & 59.3 & 61.3 & - & 9.8 & 10.5 \\
FREQ~\cite{zellers2018neural, tang2019learning} & 53.6 & 60.6 & 62.2 & 8.3 & 13.0 & 16.0 \\
UVTransE~\cite{hung2020contextual} & - & 61.2 & 64.3 & - & - & - \\
MOTIFS~\cite{zellers2018neural, tang2019learning} & 58.5 & 65.2 & 67.1 & 10.8 & 14.0 & 15.3 \\
KERN~\cite{chen2019knowledge} & - & 65.8 & 67.6 & - & 17.7 & 19.2 \\
NODIS~\cite{DBLP:journals/corr/abs-2001-04735} & 58.9 & 66.0 & 67.9 & - & - & - \\
HCNet~\cite{9084259} & 59.6 & 66.4 & 68.8 & - & - & - \\
VCTree~\cite{tang2019learning} & 60.1 & 66.4 & 68.1 & 14.0 & 17.9 & 19.4 \\
GPS-Net \cite{Lin_2020_CVPR} & 60.7 & 66.9 & 68.8 & 17.4 & 21.3 & 22.8  \\
GB-Net-$\beta$\textsuperscript{$\lozenge$} \cite{zareian2020bridging} & - & 66.6 & 68.2 & - & 22.1 & 24.0 \\
HOSE-Net~\cite{10.1145/3394171.3413575} & - & 66.7 & 69.2 & - & - & - \\
Part-Aware~\cite{tian2020part} & 61.8 & 67.7 & 69.4 & 15.2 & 19.2 & 20.9 \\
DG-PGNN \cite{Khademi_Schulte_2020} & - & 69.0 & 72.1 & - & - & - \\
\hline 
MOTIFS\textsuperscript{$\dagger$} \cite{zellers2018neural,tang2020unbiased} & 59.0 & 65.5 & 67.2 & 13.0 & 16.5 & 17.8  \\
MOTIFS-Focal\textsuperscript{$\ddagger$} \cite{lin2017focal,tang2020unbiased} & 59.2 & 65.8 & 67.7 & 10.9 & 13.9 & 15.0 \\
MOTIFS-Resample\textsuperscript{$\ddagger$} \cite{burnaev2015influence,tang2020unbiased} & 57.6 & 64.6 & 66.7 & 14.7 & 18.5 & 20.0 \\
MOTIFS-Reweight\textsuperscript{$\dagger$} & 45.4 & 54.7 & 56.5 & 14.3 & 17.3 & 18.6 \\
MOTIFS-L2+uKD\textsuperscript{$\ddagger$} \cite{DBLP:conf/bmvc/WangPL20} & 57.4 & 64.1 & 66.0 & 14.2 & 18.6 & 20.3 \\
MOTIFS-L2+cKD\textsuperscript{$\ddagger$} \cite{DBLP:conf/bmvc/WangPL20} & 57.7 & 64.6 & 66.4 & 14.4 & 18.5 & 20.2 \\
MOTIFS-TDE\textsuperscript{$\dagger$} \cite{tang2020unbiased} & 32.9 & 45.0 & 50.6 & 17.4 & 24.2 & 27.9 \\
MOTIFS-PCPL\textsuperscript{$\dagger$} \cite{yan2020pcpl} & 48.4 & 54.7 & 56.5 & 19.3 & 24.3 & 26.1 \\
MOTIFS-STL\textsuperscript{$\dagger$} \cite{chen2019soft} & 56.5 & 65.0 & 66.9 & 13.3 & 20.1 & 22.3 \\
MOTIFS-DLFE & 46.4 & 52.5 & 54.2 & 22.1 & 26.9 & 28.8 \\
\hline
VCTree\textsuperscript{$\dagger$} \cite{tang2019learning,tang2020unbiased} & 59.8 & 65.9 & 67.5 & 14.1 & 17.7 & 19.1 \\
VCTree-Reweight\textsuperscript{$\dagger$} & 53.8 & 60.7 & 62.6 & 16.3 & 19.4 & 20.4 \\
VCTree-L2+uKD\textsuperscript{$\ddagger$} \cite{DBLP:conf/bmvc/WangPL20} & 58.5 & 65.0 & 66.7 & 14.2 & 18.2 & 19.9 \\ 
VCTree-L2+cKD\textsuperscript{$\ddagger$} \cite{DBLP:conf/bmvc/WangPL20} & 59.0 & 65.4 & 67.1 & 14.4 & 18.4 & 20.0 \\ 
VCTree-TDE\textsuperscript{$\dagger$} \cite{tang2020unbiased} & 34.4 & 44.8 & 49.2 & 19.2 & 26.2 & 29.6 \\
VCTree-PCPL\textsuperscript{$\dagger$} \cite{yan2020pcpl} & 50.5 & 56.9 & 58.7 & 18.7 & 22.8 & 24.5 \\
VCTree-STL\textsuperscript{$\dagger$} \cite{chen2019soft} & 57.1 & 65.2 & 67.0 & 14.3 & 21.4 & 23.5 \\
VCTree-DLFE & 45.7 & 51.8 & 53.5 & 20.8 & 25.3 & 27.1 \\
\hline
\end{tabular}
}
\vspace{0.1em}
\caption{Recall and mean recall (with graph constraint) results in PredCls task on VG150. 
Models in the first section are with VGG backbone \cite{simonyan2014very}.
$\dagger$, $\ddagger$ and $\lozenge$ are with the same meaning as in Table 1 of the main paper.
}
% \captionsetup[table]{skip=1pt}
\label{tab:sgg_result_predcls}
% \vspace{-1em}
\end{table*}

\begin{table*}[htbp]
\centering
\scalebox{1.0}{
\begin{tabular}{l|c c c c c c}
\hline
\multicolumn{1}{c}{} & \multicolumn{6}{c}{Scene Graph Classification} \\
\hline
Model & R@20 & R@50 & R@100 & mR@20 & mR@50 & mR@100 \\ 
\hline 
IMP+~\cite{xu2017scene, chen2019knowledge} & 31.7 & 34.6 & 35.4 & - & 5.8 & 6.0 \\
FREQ~\cite{zellers2018neural, tang2019learning} & 29.3 & 32.3 & 32.9 & 5.1 & 7.2 & 8.5 \\
UVTransE~\cite{hung2020contextual} & - & 30.9 & 32.2 & - & - & - \\
MOTIFS~\cite{zellers2018neural, tang2019learning} & 32.9 & 35.8 & 36.5 & 6.3 & 7.7 & 8.2 \\
KERN~\cite{chen2019knowledge} & - & 36.7 & 37.4 & - & 9.4 & 10.0 \\
NODIS~\cite{DBLP:journals/corr/abs-2001-04735} & 36.0 & 39.8 & 40.7 & - & - & - \\
HCNet~\cite{9084259} & 34.2 & 36.6 & 37.3 & - & - & - \\
VCTree~\cite{tang2019learning} & 35.2 & 38.1 & 38.8 & 8.2 & 10.1 & 10.8 \\
GPS-Net \cite{Lin_2020_CVPR} & 36.1 & 39.2 & 40.1 & 10.0 & 11.8 & 12.6 \\
% GB-Net \cite{zareian2020bridging} & - & 19.3 & 20.9 & - & 9.6 & 10.2 & - & 6.1 & 7.3 \\
GB-Net-$\beta$\textsuperscript{$\lozenge$} \cite{zareian2020bridging} & - & 37.3 & 38.0 & - & 12.7 & 13.4 \\
HOSE-Net~\cite{10.1145/3394171.3413575} & - & 36.3 & 37.4 & - & - & - \\
Part-Aware~\cite{tian2020part} & 36.5 & 39.4 & 40.2 & 8.7 & 10.9 & 11.6 \\
DG-PGNN \cite{Khademi_Schulte_2020} & - & 39.3 & 40.1 & - & - & - \\
\hline 
MOTIFS\textsuperscript{$\dagger$} \cite{zellers2018neural,tang2020unbiased} & 36.4 & 39.5 & 40.3 & 7.2 & 8.9 & 9.4 \\
MOTIFS-Focal\textsuperscript{$\ddagger$} \cite{lin2017focal,tang2020unbiased} & 36.0 & 39.3 & 40.1 & 6.3 & 7.7 & 8.3 \\
MOTIFS-Resample\textsuperscript{$\ddagger$} \cite{burnaev2015influence,tang2020unbiased} & 34.5 & 37.9 & 38.8 & 9.1 & 11.0 & 11.8 \\
MOTIFS-Reweight\textsuperscript{$\dagger$} & 24.2 & 29.5 & 31.5 & 9.5 & 11.2 & 11.7 \\
MOTIFS-L2+uKD\textsuperscript{$\ddagger$} \cite{DBLP:conf/bmvc/WangPL20} & 35.1 & 38.5 & 39.3 & 8.6 & 10.9 & 11.8 \\
MOTIFS-L2+cKD\textsuperscript{$\ddagger$} \cite{DBLP:conf/bmvc/WangPL20} & 35.6 & 38.9 & 39.8 & 8.7 & 10.7 & 11.4 \\
MOTIFS-TDE\textsuperscript{$\dagger$} \cite{tang2020unbiased} & 21.4 & 27.1 & 29.5 & 9.9 & 13.1 & 14.9 \\
MOTIFS-PCPL\textsuperscript{$\dagger$} \cite{yan2020pcpl} & 31.9 & 35.3 & 36.1 & 9.9 & 12.0 & 12.7 \\
MOTIFS-STL\textsuperscript{$\dagger$} \cite{chen2019soft} & 35.4 & 39.9 & 40.9 & 8.5 & 12.8 & 14.1 \\
MOTIFS-DLFE & 29.0 & 32.3 & 33.1 & 12.8 & 15.2 & 15.9 \\
\hline
VCTree\textsuperscript{$\dagger$} \cite{tang2019learning,tang2020unbiased} & 42.1 & 45.8 & 46.8 & 9.1 & 11.3 & 12.0 \\
VCTree-Reweight\textsuperscript{$\dagger$} & 38.0 & 42.3 & 43.5 & 10.6 & 12.5 & 13.1 \\
VCTree-L2+uKD\textsuperscript{$\ddagger$} \cite{DBLP:conf/bmvc/WangPL20} & 40.9 & 44.7 & 45.6 & 9.9 & 12.4 & 13.4 \\ 
VCTree-L2+cKD\textsuperscript{$\ddagger$} \cite{DBLP:conf/bmvc/WangPL20} & 41.4 & 45.2 & 46.1 & 9.7 & 12.4 & 13.1 \\ 
VCTree-TDE\textsuperscript{$\dagger$} \cite{tang2020unbiased} & 21.7 & 28..8 & 32.0 & 11.2 & 15.2 & 17.5 \\
VCTree-PCPL\textsuperscript{$\dagger$} \cite{yan2020pcpl} & 36.5 & 40.6 & 41.7 & 12.7 & 15.2 & 16.1 \\
VCTree-STL\textsuperscript{$\dagger$} \cite{chen2019soft} & 40.6 & 45.7 & 46.9 & 10.5 & 14.6 & 16.6 \\
VCTree-DLFE & 29.7 & 33.5 & 34.6 & 15.8 & 18.9 & 20.0 \\
\hline
\end{tabular}
}
\vspace{0.1em}
\caption{Recall and mean recall (with graph constraint) results in SGCls task on VG150. 
Models in the first section are with VGG backbone \cite{simonyan2014very}.
$\dagger$, $\ddagger$ and $\lozenge$ are with the same meaning as in Table 1 of the main paper.
}
% \captionsetup[table]{skip=1pt}
\label{tab:sgg_result_sgcls}
% \vspace{-1em}
\end{table*}

\begin{table*}[htbp]
\centering
\scalebox{1.0}{
\begin{tabular}{l |c c c c c c}
\hline
\multicolumn{1}{c}{} & \multicolumn{6}{c}{Scene Graph Detection} \\
\hline
Model & R@20 & R@50 & R@100 & mR@20 & mR@50 & mR@100 \\ 
\hline 
IMP+~\cite{xu2017scene, chen2019knowledge} & 14.6 & 20.7 & 24.5 & - & 3.8 & 4.8 \\
FREQ~\cite{zellers2018neural, tang2019learning} & 20.1 & 26.2 & 30.1 & 4.5 & 6.1 & 7.1 \\
UVTransE~\cite{hung2020contextual} & - & 25.3 & 28.5 & - & - & - \\
MOTIFS~\cite{zellers2018neural, tang2019learning} & 21.4 & 27.2 & 30.3 & 4.2 & 5.7 & 6.6  \\
KERN~\cite{chen2019knowledge} & - & 27.1 & 29.8 & - & 6.4 & 7.3  \\
NODIS~\cite{DBLP:journals/corr/abs-2001-04735} & 21.5 & 27.4 & 30.7 & - & - & - \\
HCNet~\cite{9084259} & 22.6 & 28.0 & 31.2 & - & - & - \\
VCTree~\cite{tang2019learning} & 22.0 & 27.9 & 31.3 & 5.2 & 6.9 & 8.0 \\
GPS-Net \cite{Lin_2020_CVPR} & 22.6 & 28.4 & 31.7 & 6.9 & 8.7 & 9.8 \\
GB-Net-$\beta$\textsuperscript{$\lozenge$} \cite{zareian2020bridging} & - & 26.3 & 29.9 & - & 7.1 & 8.5 \\
HOSE-Net~\cite{10.1145/3394171.3413575} & - & 28.9 & 33.3 & - & - & - \\
Part-Aware~\cite{tian2020part} & 23.4 & 29.4 & 32.7 & 5.7 & 7.7 & 8.8 \\
DG-PGNN \cite{Khademi_Schulte_2020} & - & 31.2 & 32.5 & - & - & - \\
\hline 
MOTIFS\textsuperscript{$\dagger$} \cite{zellers2018neural,tang2020unbiased} & 25.8 & 33.1 & 37.6 & 5.3 & 7.3 & 8.6 \\
MOTIFS-Focal\textsuperscript{$\ddagger$} \cite{lin2017focal,tang2020unbiased} & 24.7 & 31.7 & 36.7 & 3.9 & 5.3 & 6.6 \\
MOTIFS-Resample\textsuperscript{$\ddagger$} \cite{burnaev2015influence,tang2020unbiased} & 23.2 & 30.5 & 35.4 & 5.9 & 8.2 & 9.7 \\
MOTIFS-Reweight\textsuperscript{$\dagger$} & 18.3 & 24.4 & 29.3 & 6.7 & 9.2 & 10.9 \\
MOTIFS-L2+uKD\textsuperscript{$\ddagger$} \cite{DBLP:conf/bmvc/WangPL20} & 24.8 & 32.2 & 36.8 & 5.7 & 7.9 & 9.5 \\
MOTIFS-L2+cKD\textsuperscript{$\ddagger$} \cite{DBLP:conf/bmvc/WangPL20} & 25.2 & 32.5 & 37.1 & 5.8 & 8.1 & 9.6 \\
MOTIFS-TDE\textsuperscript{$\dagger$} \cite{tang2020unbiased} & 12.4 & 17.3 & 20.8 & 6.7 & 9.2 & 11.1 \\
MOTIFS-PCPL\textsuperscript{$\dagger$} \cite{yan2020pcpl} & 21.3 & 27.8 & 31.7 & 8.0 & 10.7 & 12.6 \\
MOTIFS-STL\textsuperscript{$\dagger$} \cite{chen2019soft} & 22.5 & 29.9 & 34.9 & 5.4 & 7.6 & 9.1 \\
MOTIFS-DLFE & 18.9 & 25.4 & 29.4 & 8.6 & 11.7 & 13.8 \\
\hline
VCTree\textsuperscript{$\dagger$} \cite{tang2019learning,tang2020unbiased} & 24.1 & 30.8 & 35.2 & 5.2 & 7.1 & 8.3 \\
VCTree-Reweight\textsuperscript{$\dagger$} & 20.8 & 27.8 & 32.0 & 6.6 & 8.7 & 10.1\\
VCTree-L2+uKD\textsuperscript{$\ddagger$} \cite{DBLP:conf/bmvc/WangPL20} & 24.4 & 31.6 & 35.9 & 5.7 & 7.7 & 9.2 \\ 
VCTree-L2+cKD\textsuperscript{$\ddagger$} \cite{DBLP:conf/bmvc/WangPL20} & 24.8 & 32.0 & 36.1 & 5.7 & 7.7 & 9.1 \\ 
VCTree-TDE\textsuperscript{$\dagger$} \cite{tang2020unbiased} & 12.3 & 17.3 & 20.9 & 6.8 & 9.5 & 11.4 \\
VCTree-PCPL\textsuperscript{$\dagger$} \cite{yan2020pcpl} & 20.5 & 26.6 & 30.3 & 8.1 & 10.8 & 12.6 \\
VCTree-STL\textsuperscript{$\dagger$} \cite{chen2019soft} & 21.6 & 28.8 & 33.6 & 5.1 & 7.1 & 8.4 \\
VCTree-DLFE & 16.8 & 22.7 & 26.3 & 8.6 & 11.8 & 13.8 \\
\hline
\end{tabular}
}
\vspace{0.1em}
\caption{Recall and mean recall (with graph constraint) results in SGDet task on VG150. 
Models in the first section are with VGG backbone \cite{simonyan2014very}.
$\dagger$, $\ddagger$ and $\lozenge$ are with the same meaning as in Table 1 of the main paper.
}
% \captionsetup[table]{skip=1pt}
\label{tab:sgg_result_sgdet}
% \vspace{-1em}
\end{table*}

\section{Additional Qualitative Results}
\label{sec:appendix_qualitative_results}
\begin{figure*}[t!]
\begin{center}
\includegraphics[width=\textwidth]{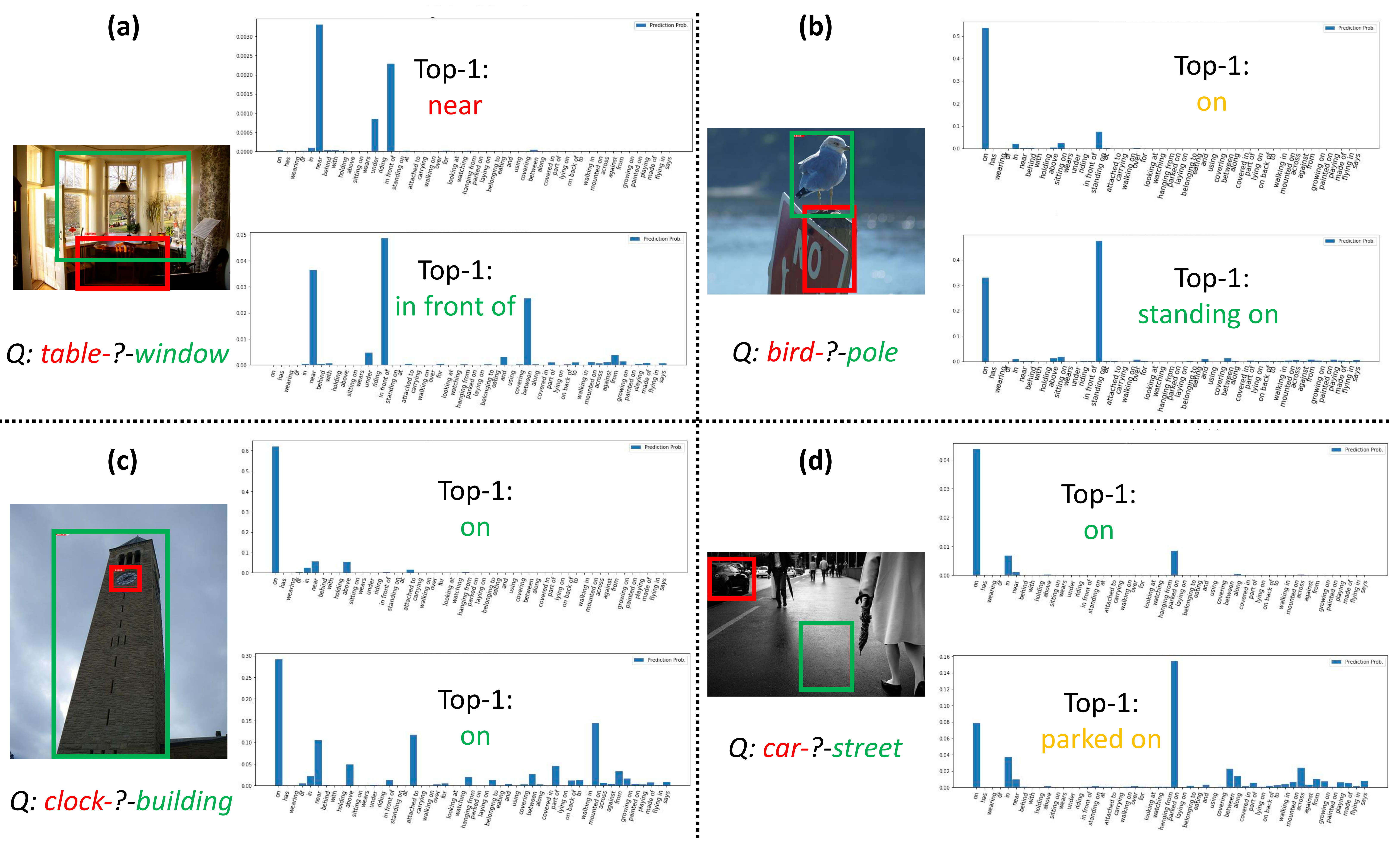}
\end{center}
\vspace{-1em}
  \caption{Confidence distributions over the predicates, produced by MOTIFS (top of each example) and MOTIFS with DLFE (bottom). \textcolor{Green}{Green}, \textcolor{Red}{red} and \textcolor{Dandelion}{tangerine} predicates denote correct (GT), incorrect (Non-GT and weird) and acceptable (Non-GT but reasonable), respectively.}
\label{fig:sgg_vis_dist2}
\vspace{-1em}
\end{figure*}

We present additional results on the change of confidence distribution after applying DLFE to MOTIFS in Figure \ref{fig:sgg_vis_dist2}. 
For {\fontfamily{qcr}\selectfont table-?-window} in example (a), instead of the prediction of {\fontfamily{qcr}\selectfont near} by MOTIFS, MOTIFS-DLFE predicts a more descriptive {\fontfamily{qcr}\selectfont in front of} which matches the ground truth.
The same applies to {\fontfamily{qcr}\selectfont bird-?-pole} in example (b) where MOTIFS-DLFE's {\fontfamily{qcr}\selectfont standing on} is better than MOTIFS's {\fontfamily{qcr}\selectfont on}. 
(c) is an interesting example that, while the top prediction {\fontfamily{qcr}\selectfont on} by MOTIFS-DLFE is the same as that of MOTIFS, more descriptive predicates such as {\fontfamily{qcr}\selectfont mounted on}, {\fontfamily{qcr}\selectfont attached to} are assigned with the higher confidence scores by MOTIFS-DLFE.
This fact should make it easier for, \emph{e.g.}, non-graph constraint mean recall (ng-mR@$K$), to \textit{recall} these fine-grained predicates
Finally, {\fontfamily{qcr}\selectfont car-?-street} is another example that MOTIFS-DLFE produces more descriptive {\fontfamily{qcr}\selectfont parked on} rather than {\fontfamily{qcr}\selectfont on}; however, due to the missing labels {\fontfamily{qcr}\selectfont parked on} is not in the ground truth.

\end{document}